\documentclass[10pt]{article} 
\newcommand{\ie}{i.e.,\ }
\usepackage[preprint]{tmlr}

\usepackage{amsmath,amsfonts,bm}

\def\eqref#1{equation~\ref{#1}}

\def\1{\bm{1}}

\DeclareMathAlphabet{\mathsfit}{\encodingdefault}{\sfdefault}{m}{sl}
\SetMathAlphabet{\mathsfit}{bold}{\encodingdefault}{\sfdefault}{bx}{n}

\usepackage{hyperref}
\usepackage{url}
\usepackage{microtype}
\usepackage{graphicx}
\usepackage{subcaption}
\usepackage{fp}
\usepackage{subcaption}
\usepackage{comment}
\usepackage{subcaption}
\usepackage{float}
\usepackage{booktabs}
\usepackage{hyperref}
\usepackage{amsmath}
\usepackage{arydshln} 
\usepackage{subcaption}
\usepackage{amssymb}
\usepackage{xfp}
\usepackage{float} 
\usepackage{placeins}
\usepackage{mathtools}
\usepackage{amsthm}
\let\AND\relax
\usepackage{algorithm}
\usepackage{algorithmic}
\usepackage{listings}
\usepackage{multirow}
\usepackage{xfp}
\usepackage{colortbl}
\usepackage{xcolor}

\usepackage{multirow}
\usepackage[normalem]{ulem}
\lstdefinestyle{cppstyle}{
  language=C++,
  basicstyle=\ttfamily\footnotesize,
  numbers=left,
  numberstyle=\tiny,
  stepnumber=1,
  numbersep=5pt,
  showstringspaces=false,
  tabsize=2,
  breaklines=true,
  breakatwhitespace=false,
  escapeinside={\%*}{*)},
  frame=single,
  xleftmargin=2pt,
  xrightmargin=2pt
}

\renewcommand{\algorithmiccomment}[1]{\hfill$\triangleright$ #1}

\makeatletter
\newcommand{\WHILECOMMENT}[2]{%
  \ALC@it\algorithmicwhile\ #1\ \algorithmicdo\ \algorithmiccomment{#2}%
  \begin{ALC@whl}%
}
\makeatother

\newcommand{\calcpercent}[2]{%
    \FPeval{\result}{round(#1/#2*100, 1)}%
    \result\%%
}

\usepackage{listings}

\usepackage{comment}
\usepackage{tabularx}
\usepackage{makecell}

\newcommand{\eat}[1]{}
\usepackage[capitalize,noabbrev]{cleveref}

\title{Steered Generation via Gradient-Based Optimization on Sparse Query Features}

\author{\name Sumanta Bhattacharyya \email sbhatt54@uic.edu  \\
      \addr Department of Computer Science\\ University of Illinois Chicago \\\\
      \AND
      \name Pedram Rooshenas \email pedram@uic.edu \\
      \addr  Department of Computer Science\\ University of Illinois Chicago
      }

\def\month{MM}  
\def\year{YYYY} 
\def\openreview{\url{https://openreview.net/forum?id=XXXX}} 

\begin{document}

\maketitle

\begin{abstract}
Latent steering exploits internal representations of Large Language Models (LLMs) to guide generation, yet interventions on dense states can entangle distinct semantic features. In this paper, we investigate attention query activations as a high-fidelity site for precise control, hypothesizing that manipulating the attention mechanism itself offers sharper steerability than general state interventions. We introduce Prototype-Based Sparse Steering, a framework that applies Sparse Autoencoders (SAEs) specifically to query activations, to decompose them into interpretable features, then apply gradient-based optimization during inference to align the sparse representation with class prototypes of target behaviors. To validate this architectural insight, we first analyze the mechanism in Textualized Gridworld, a controlled environment for verifiable planning constraints. We demonstrate that optimizing sparse query features enables effective navigation of rigid planning requirements (\ie safe vs. short paths), confirming the method's ability to satisfy objective rules. We then demonstrate the framework's versatility by training SAEs on a high-dimensional educational domain, where the framework steers the cognitive complexity of feedback (\ie Bloom’s Taxonomy). Our experiments establish that sparse query representations provide the necessary disentanglement for unified, interpretable control over both logical planning and stylistic nuance.
\end{abstract}
\section{Introduction}

LLMs have emerged as the de facto natural language interface for modern AI systems. However, deploying these models in real-world scenarios requires more than fluent generation; it necessitates precise adherence to user-defined requirements regarding style, safety, and logical consistency. Consequently, research on controlled text generation in LLMs has been extensively explored for general-purpose applications (see the comprehensive survey by~\citet{ctg-survey}). Unlike standard text generation, controlled text generation~\citep{mireshghallah2022mix} guides LLMs to generate outputs that adhere to specific constraints.

To achieve this control, prompt engineering has become a prevalent strategy. However, controlling text generation solely through prompting can be ineffective~\citep{ye2022unreliability}, as prompts often require extensive contextual information to generate relevant responses. When additional instructions for controlling style and nuance are included, they may be diluted or ignored within long prompt contexts~\citep{liu2023lost}. Moreover, open-source and smaller LLMs provide limited control over generated text when relying solely on prompt-based methods~\citep{salinas2024butterfly}. In practice, this gap is most visible in smaller open-source models: the same instruction that yields stable stylistic adherence in frontier models can produce inconsistent or diluted control in smaller models. A common mitigation is to include a few labeled demonstrations in the prompt (in-context learning) to anchor the desired attribute~\citep{brown2020language}. However, increasing the number of demonstrations leads to significant memory consumption and computational overhead, particularly with the larger context windows needed for complex tasks~\citep{wei2022chain,lilong}. These limitations motivate approaches that do not rely on expanding the prompt, but instead intervene directly in the model's internal representations at inference time.

\noindent Latent-space steering methods, often referred to as activation engineering, enable controlled text generation by intervening directly in the internal activations of large language models during inference~\citep{Dathathri_Madotto_Lan_Hung_Frank_Molino_Yosinski_Liu_2019,Liu_Sap_Lu_Swayamdipta_Bhagavatula_Smith_Choi_2021,Khalifa_ElSahar_Dymetman_2020}. These methods typically use steering vectors, which are directions in activation space associated with semantic, behavioral, or stylistic attributes, to shift model behavior in predictable ways~\citep{hernandez2023inspecting,sun2024massive}. Compared with in-context prompting, activation engineering provides a more direct mechanism for controlling attributes such as style, tone, formality, or factuality, without expanding the prompt context. However, dense activation spaces can entangle multiple latent factors, consistent with the superposition hypothesis~\citep{elhage2022toy}, making interventions less precise and more prone to unintended side effects. Sparse representations, including those learned by Sparse Auto-Encoders (SAEs; \citealp{Makhzani2013kSparseA}), offer a natural way to factorize control directions by encouraging updates to concentrate on a smaller set of features~\citep{cunningham2023sparse,bricken2023towards}. This sparsity can support more isolated and independent latent updates, enabling more precise and reliable control over target attributes during generation. 

\noindent Prior steering largely adds static vectors to the residual stream or attention-head outputs, which can overwrite context and entangle attributes; we instead ask whether a more local site, optimized rather than fixed, yields cleaner control. In this paper, we hypothesize that attention query activations serve as a high-fidelity control surface for generation, as they encode the model's intent to retrieve information from context. Because queries parameterize how tokens attend to and retrieve information, small and localized query perturbations can modify what the model uses while substantially reducing global distortion of the residual stream. We propose an approach for controlled generation that combines SAEs with gradient-based latent search. Inspired by prototypical networks~\citep{snell2017prototypical}, originally designed for few-shot learning, we represent target attributes as prototype distributions in the sparse latent space. Instead of adding a static steering vector, we treat control as an optimization problem, steering encoded features to increase alignment with the desired prototype distribution.



We validate our approach across complementary control settings. For hard, verifiable control, we evaluate on Textualized Gridworld (TGW;~\citealp{kim2024language}), adapted from classic RL planning setups~\citep{zai2020deep}, where generated paths must satisfy explicit safety and length constraints. This setting enables exact, rule-based evaluation: we can directly measure whether the generated path remains valid while satisfying the target objective, such as producing a safe, short, or long path. For soft semantic control, we transfer our method to an educational feedback setting and steer Bloom-level cognitive complexity~\citep{blooms}, demonstrating control over nuanced pedagogical and stylistic attributes. Finally, we evaluate on TruthfulQA~\citep{lin2022truthfulqa} as an external benchmark to assess whether gradient-based steering improves truthfulness and informativeness while reducing distributional distortion compared with static vector-addition baselines.

Our primary contributions and findings are the following:

1) \textbf{Prototype-Based Sparse Query Optimization:} We formulate inference-time control as gradient-based optimization over sparse query latents, using a prototype-distribution objective defined by a softmax over negative distances to target attribute prototypes. This enables adaptive steering toward semantic or behavioral targets while keeping the base model weights fixed.

2) \textbf{Query-Level Control Fidelity:} We provide quantitative evidence that attention query activations offer a more local and lower-distortion intervention site than the residual stream. Specifically, query-level interventions exhibit bounded cross-layer activation deviation and lower next-token Jensen--Shannon divergence at matched steering strength, supporting their use for both planning and stylistic control. Recent independent work on query-level steering~\citep{torop2025disco} provides complementary evidence for the same broader observation.

3) \textbf{Steering-Oriented Evaluation Data:} We construct a new Bloom's Taxonomy dataset for cognitive style steering~\citep{blooms} and derive a filtered Gridworld steering subset from Textualized Gridworld~\citep{kim2024language}. The Gridworld subset is designed to contain mutually distinct safe, short, and long paths, enabling rule-based evaluation of hard steering constraints.

4) \textbf{Experimental Validation Across Steering Regimes:} We evaluate our method on hard planning constraints, soft cognitive-style control, and TruthfulQA truthfulness steering. Across TGW experiments on three open-weight models, Bloom-level steering, and TruthfulQA, our method matches or exceeds prompting, residual-stream steering, attention-head steering, dense query steering, and static sparse steering baselines, while often inducing substantially lower distributional cost.

\section{Proposed Steering Framework}

\begin{figure*}[h]
    \centering
    \includegraphics[width=0.9\linewidth]{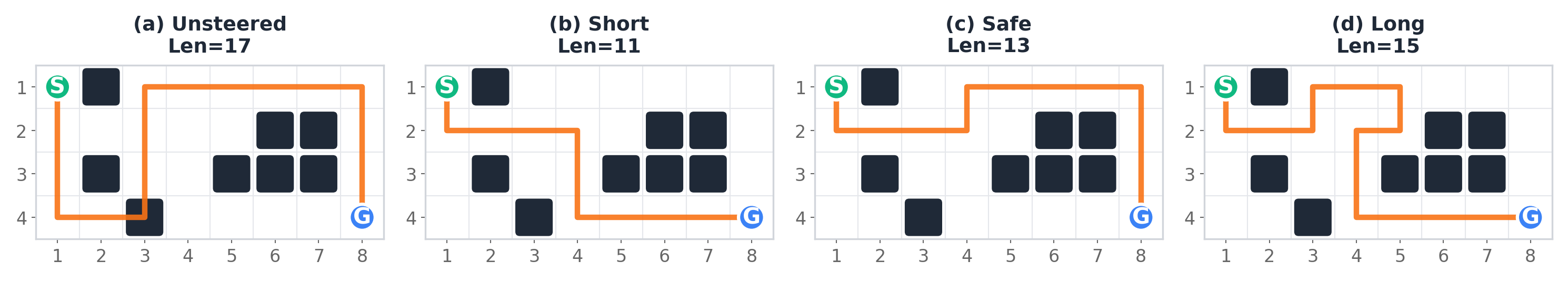}
    \caption{Quantifiable steering in TGW. We steer generation to short, safe (lower wall-adjacency score), and long targets; success is verifiable via path validity and attribute metrics. The unsteered output may be invalid.}
    \label{fig:tgw_steer}
\end{figure*}

We study inference-time steering for controlled generation: given an input prompt and a target attribute, the goal is to bias an LLM's output toward that attribute while preserving task-relevant content and coherence.
For example, in Textualized Gridworld (TGW) a prompt specifies a grid map with walls, a start cell, and a goal cell, and the model must output a sequence of cells representing a path from start to goal~\citep{kim2024language}. We define a target attribute to require the path be safe,  short, or long. In our TGW setting, we quantify safety by the number of wall-adjacent cells along the path, so safer paths have lower wall-adjacency scores; steering succeeds only if the generated cell sequence is a valid path (each step moves to an adjacent cell and does not enter wall cells) while satisfying the specified attribute.
TGW illustrates a hard, verifiable regime: preservation corresponds to maintaining structural validity with respect to the task (a well-formed path), while attributes such as safety and length are directly measurable on the resulting path. Figure~\ref{fig:tgw_steer} shows an example of steering a single base model output toward different target attributes. In TGW, the steered output is exactly verifiable: we can check path validity and category-specific optimality by computing the path score (e.g., wall-adjacency count for safety or path length for short/long targets) and comparing it to the score of a reference (gold) solution.

Achieving these verifiable targets requires modifying the model's internal trajectory at inference time, which motivates interventions on intermediate activations rather than relying solely on prompting. Dense internal activations provide a valid intervention point for inference-time steering, but updates in this space are inherently coupled across many latent factors, making iterative optimization prone to interference. To improve the conditioning of the control space without committing to a specific intervention site, we reparameterize the steerable activation using a Sparse Auto-Encoder (SAE). The SAE enforces sparsity through an $L_1$ constraint on latent features, encouraging updates to activate only a small subset of directions~\citep{konda2014zero,lee2007sparse}. In our setting, this sparsity serves to factorize control directions, enabling more isolated and targeted updates during inference-time optimization.


Let $s = \mathrm{LLM}_{\leq \ell}(x, y)$ denote the activation extracted at intervention layer $\ell$ (we analyze the choice of site in Section~\ref{subsec:qli}) for the full sequence consisting of the prompt $x$ and generated response $y$. The SAE encoder maps this activation to a sparse latent representation $z = \mathrm{Enc}(s)$, and the decoder reconstructs it back to the original activation space as $\hat{s} = \mathrm{Dec}(z)$. Steering is performed by optimizing $z$ while keeping all LLM and SAE parameters frozen. The SAE parameters are shared across token positions, and the latent representation is optimized jointly over the full sequence. After optimization, the reconstructed activation $\hat{s}$ is injected back at the same intervention point, and generation continues normally.

\begin{algorithm}
\caption{Attribute-conditioned Steering using Sparse Autoencoder}
\label{alg:steering}
\begin{small}
\begin{algorithmic}[1]

\STATE \textbf{Input:} Source text $x$, target attribute $k_T$, Number of attribute class $K$, support sets $D_{k}$,
$\mathrm{LLM}$, Enc, Dec, step size $\eta$, threshold $\epsilon$, intervention layer $\ell$
\STATE \textbf{Output:} Steered text toward target attribute 

\FOR{$k \in \{1,\ldots,K\}$} 
     \STATE $c_k \gets \frac{1}{|D_k|} \sum_{((x_i, y_i), a_i)\in D_k} 
     \mathrm{Enc} (\mathrm{LLM}_{\leq \ell}(x_i,y_i))$ \COMMENT{Calculate the center for each prototype}
\ENDFOR
\STATE $y' \gets \mathrm{LLM}(x)$ \COMMENT{Get the initial model output}
\STATE $t \gets 0$; $z^{(t)} \gets \text{Enc}(\mathrm{LLM}_{\leq \ell}(x_i,y'))$ \COMMENT{Obtain the initial latent representation}
\WHILECOMMENT{$\left\lVert \nabla_z \log P_\phi(a=k_T \mid z^{(t)})\right\rVert_2^2 > \epsilon$ } {Continue until the steering update converges}
    \FOR{$k \in \{1,\ldots,K\}$ } 
        \STATE $d_k \gets \lVert z^{(t)} - c_k \rVert_2^2$ \COMMENT{Find the distance of latent representation to each prototype center}
    \ENDFOR
    \STATE $P_\phi(a=k_T \mid z^{(t)}) \gets
    \frac{\exp(-d_{k_T})}{\sum_{j=1}^{K} \exp(-d_j)}$ \COMMENT{Calc. the likelihood of the target prototype using negative distance}
    \STATE $z^{(t+1)} \gets z^{(t)} + \eta \nabla_z \log P_\phi(a=k_T \mid z^{(t)})$ \COMMENT{Gradient step to increase the likelihood of the target prototype}
    \STATE $t \gets t + 1$
\ENDWHILE

\STATE \textbf{return} $\mathrm{LLM}_{\geq \ell}(\text{Dec}(z^{(t)}))$

\end{algorithmic}
\end{small}
\end{algorithm}

\subsection{Attribute-Conditioned Steering Objective}

To define steering targets in the sparse latent space learned by the SAE, we borrow the idea of prototype-based representations~\citep{snell2017prototypical}. Rather than using prototypes for classification, we use them to define target regions in the latent space that guide inference-time control.

For each attribute class $k$, we define a support set $D_k$ consisting of training examples labeled with attribute $k$: $D_k = \{((x_i, y_i), a_i) : a_i = k\}$, where $x_i$ is the prompt, $y_i$ is the response, and $a_i$ is the attribute label. For each class, we define the prototype $c_k$ as the mean sparse latent representation of the support examples:
\begin{equation}
c_k = \frac{1}{|D_k|} \sum_{((x_i, y_i), a_i)\in D_k} \mathrm{Enc}(\mathrm{LLM}_{\leq \ell}(x_i,y_i)).
\end{equation}

Instead of directly projecting the latent representation onto the target prototype $c_k$, which empirically leads to reduced diversity in generated outputs (see Section~\ref{sec:tcd}), we perform gradient-based optimization so that alignment is achieved gradually while respecting the latent structure.

We define a metric-based distribution over attribute classes using a softmax over negative distances to the corresponding prototypes:
\begin{equation}
P_\phi(a=k_T \mid z) =
    \frac{\exp(-d_{k_T})}{\sum_{j=1}^{K} \exp(-d_j)},
\end{equation}
where $k_T$ is the target attribute class, and $d_k$ is the distance of latent representation $z$ from the center of prototype $k$ using a fixed distance metric (squared Euclidean distance in our experiments).
This probability serves as a continuous control score measuring alignment between the current generation and the target attribute, rather than as a classifier.

Formally, given a generated response $y'$ to prompt $x$, steering begins by extracting the activation at intervention layer $\ell$ and encoding it into an initial sparse latent representation,
\(
z^{(0)} = \mathrm{Enc}\!\left(\mathrm{LLM}_{\leq \ell}(x,y')\right).
\)
We then optimize $z$ directly to increase its alignment with the target attribute $k_T$. Concretely, we maximize the target log-probability $\log P_\phi(a = k_T \mid z)$, 
where $P_\phi$ is defined via distances to class prototypes in the sparse latent space.
Optimization proceeds via gradient ascent:
\begin{equation}
z^{(t+1)} = z^{(t)} + \eta \nabla_z \log P_\phi(a = k_T \mid z^{(t)}),
\label{eq:steering_update}
\end{equation}
where $\eta$ controls the steering strength. 

After convergence, the optimized latent is decoded back to the activation space using $\text{Dec}$, and generation continues normally. Algorithm~\ref{alg:steering} illustrates the full steering procedure.

\subsection{Query-Level Intervention}
\label{subsec:qli}

Prior work intervenes at different architectural sites and uses either static steering vectors or state-dependent signals~\citep{hernandez2023inspecting,sun2024massive,li2023inference,wang2025semantics,li2025fairsteer}. However, the reliability of these methods is limited by geometric and separability constraints~\citep{im2025unified}. In particular, when generation must satisfy hard constraints, such as logical or structural validity, interventions that broadly perturb dense representations can overwrite context and induce violations. Under soft semantic control, they can drift or inadvertently blend attributes, yielding unintended attribute mixing. Consistent with this, prior residual-stream steering methods report qualitative degradation and side effects at higher intervention strengths~\citep{turner2023activation,stolfo2024improving,elhage2021mathematical}.

To preserve high-fidelity locality, we prioritize intervening on self-attention query activations rather than the residual stream. Queries provide a more local control handle because they modulate what information is retrieved through attention without directly overwriting the accumulated residual state; specifically, they affect the attention weights, so each head produces a convex combination of value embeddings whose output lies in the convex hull of the values rather than an unconstrained additive perturbation.

\begin{figure*}[t]
    \centering
    \begin{subfigure}[t]{0.40\textwidth}
        \centering
        \includegraphics[width=\linewidth, height=6.0cm, keepaspectratio]{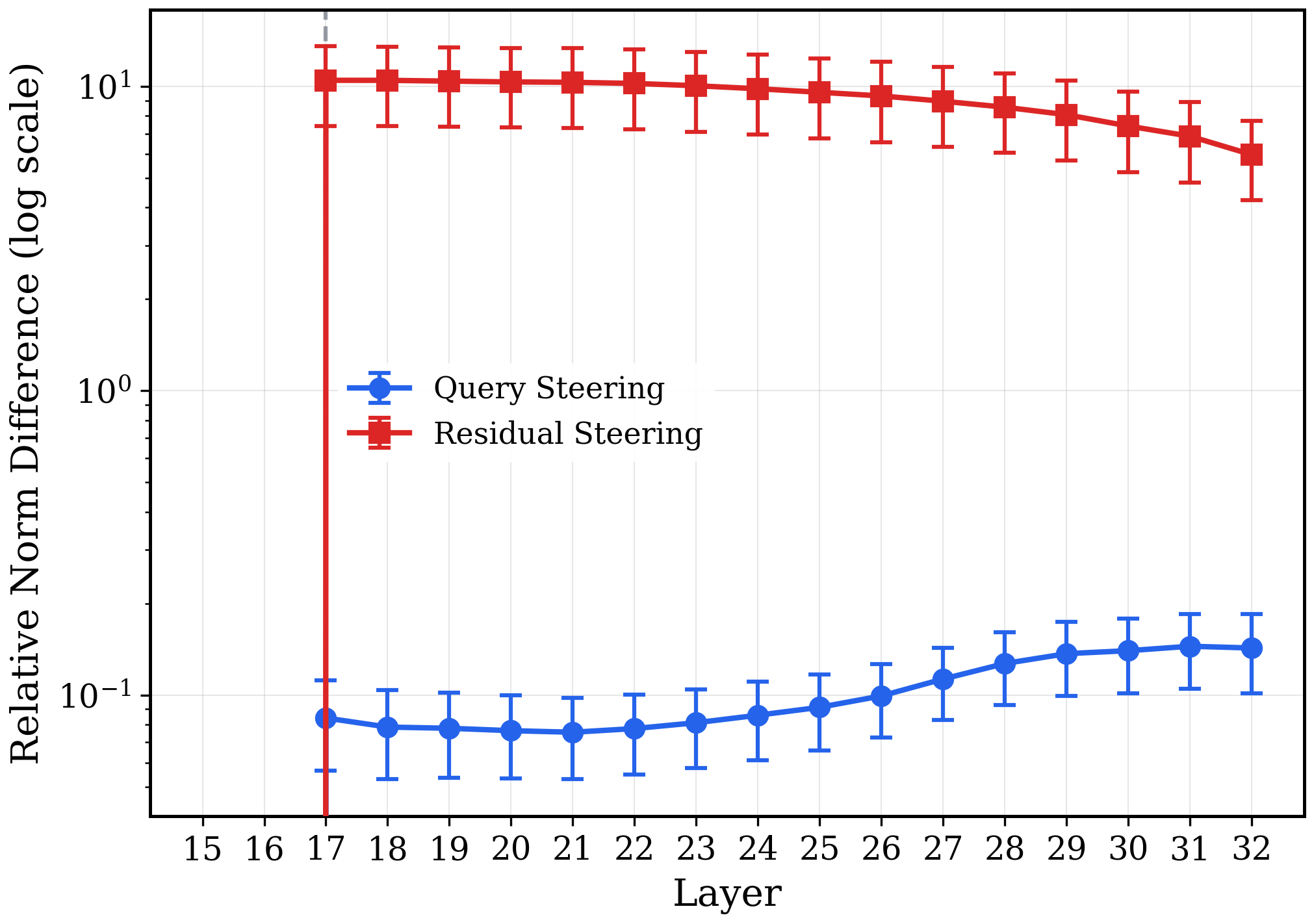}
        \caption{Relative activation deviation $\delta_l$ (log scale) across layers after applying a fixed-strength intervention at layer 17. Query-level steering remains bounded and slowly varying, whereas residual-stream steering induces one to two orders of magnitude larger, persistent deviation (error bars show variability across samples).}
        \label{fig:decay_main}
    \end{subfigure}
    \hfill
    \begin{subfigure}[t]{0.50\textwidth}
        \centering
        \includegraphics[width=\linewidth, height=6.0cm, keepaspectratio]{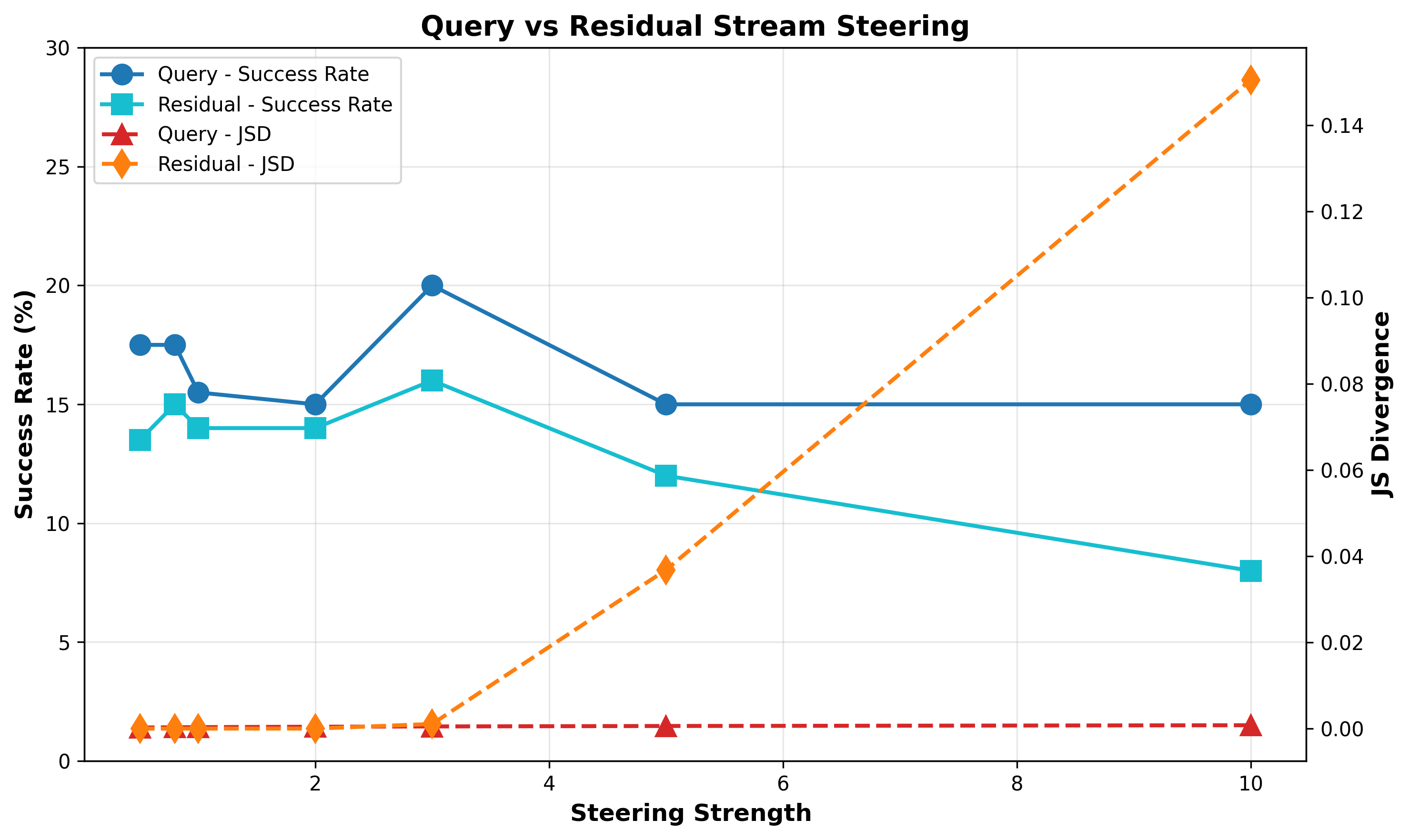}
        \caption{Query vs.\ residual stream steering tradeoffs. Task success rate (left axis) and Jensen Shannon divergence (JSD; right axis) as a function of steering strength $\eta$ for query-based (attention) and residual stream steering.}
        \label{fig:pareto_main}
    \end{subfigure}
    \caption{Quantitative diagnostics for intervention locality. (a) Propagation of activation deviation across layers. (b) Task success vs.\ distributional distortion trade-off.}
    \label{fig:locality_diagnostics}
\end{figure*}

To justify this design choice, we run two locality diagnostics (these are not task-performance evaluations, which we defer to Sec.~\ref{sec:exp}) comparing interventions (query steering and residual steering) applied at the same layer. First, we measure how an intervention propagates through each layer $\ell$ above the intervention site by tracking the relative change in activations compared to an unmodified forward pass:
\begin{align}
\delta_\ell = \frac{1}{T} \sum_{t=1}^{T} 
\frac{\| \mathbf{h}_{\mathrm{steered}}^{(\ell,t)} - \mathbf{h}_{\mathrm{base}}^{(\ell,t)} \|_2}
{\| \mathbf{h}_{\mathrm{base}}^{(\ell,t)} \|_2},
\end{align}
where $T$ is the sequence length and $\mathbf{h}^{(\ell,t)}$ is the activation vector at layer $\ell$ and token position $t$. As shown in Figure~\ref{fig:decay_main}, query interventions yield a bounded, slowly varying deviation across layers, while residual-stream interventions induce a large, persistent deviation, consistent with a more disruptive overwrite of local context.

Second, beyond activation propagation, we characterize the accuracy-distortion trade-off induced by the intervention site as steering strength increases. The core question is whether stronger steering yields commensurate gains in task success, or instead primarily manifests as unintended shifts in the model's output distribution. We quantify this distributional cost using the Jensen--Shannon divergence (JSD) between the baseline and steered next-token distributions. Let $P_{b}$ and $P_{s}$ denote the probability distributions over the vocabulary for the baseline and steered models, respectively:
\begin{equation}
    \text{JSD}(P_{b} \parallel P_{s}) = \frac{1}{2} D_{\text{KL}}(P_{b} \parallel M) + \frac{1}{2} D_{\text{KL}}(P_{s} \parallel M),
\end{equation}
where $M = \frac{1}{2}(P_{b} + P_{s})$ is the mixture distribution. We sweep steering strengths ($\eta$) and plot task success against JSD cost. Figure~\ref{fig:pareto_main} provides a quantitative confirmation of the trade-off: residual-stream steering exhibits rapidly increasing distortion with diminishing returns in task success, whereas query-level steering achieves comparable control with substantially lower distributional cost. 

While query perturbations can still propagate through downstream layers, our locality diagnostics, including bounded activation deviation $\delta_\ell$ and lower next-token JSD at matched steering strength, indicate that query-level interventions induce systematically smaller distributional shifts than residual-stream additions in our settings. Recent independent work reaches a related conclusion: \citet{torop2025disco} report similar benefits from query- and value-level interventions. Our diagnostics complement theirs by directly quantifying the distortion advantage of query-level steering. Together, these results motivate query activations as a suitable intervention site for reliable steering under both hard and soft constraint regimes.





\subsection{Training Sparse Autoenoders}
Our SAE architecture is a single-layer ReLU autoencoder (similar to~\citet{bricken2023towards}).
For activation $s$ at an intervention site, we define a shallow autoencoder with encoder
$\mathrm{Enc}(s) = \mathrm{ReLU}(W_e s + b_e)$ and decoder $\mathrm{Dec}(z) = \tilde{W}_d\, z$,
where $(W_e, b_e)$ and $W_d$ are the encoder and decoder parameters, respectively, and
$z = \mathrm{Enc}(s)$ is the encoded latent representation. To prevent the $L_1$ penalty from
being trivially minimized by rescaling, each decoder column (i.e., the dictionary direction for
latent $i$) is normalized to unit norm,
\[
\tilde{W}_d^{(i)} = \frac{W_d^{(i)}}{\max\!\left(\|W_d^{(i)}\|_2,\, \gamma\right)},
\]
where $W_d^{(i)}$ denotes the $i$-th column of $W_d$ and $\gamma$ is a small constant for
numerical stability. We then define the training loss as a combination of a reconstruction loss,
a sparsity penalty, and a bias decay term:
\begin{align}
\label{eq:sae_loss}
\mathcal{L}(s) = \underbrace{\|\hat s - s\|_2^2}_{\text{Reconstruction loss}}
+ \underbrace{\lambda\|z\|_1}_{\text{Sparsity loss}}
+ \underbrace{\beta\|b_e\|_2^2}_{\text{Bias decay}},
\end{align}
where $\hat s = \mathrm{Dec}(z)$ is the reconstructed activation, $\lambda$ controls the $L_1$
sparsity strength, and $\beta$ controls the bias decay.





We train one SAE per query attention head. Because each head in the multi-head attention mechanism operates independently and captures different aspects of the input, we train these SAEs separately rather than sharing parameters across heads. Each SAE is trained on the query activations of its corresponding head, extracted from the self-attention layer at the intervention site.


steering.

\eat{This gradient is then used to guide the steering process toward the desired class:
\begin{align}
z_{i \rightarrow j} = z_{i} + \eta\frac{\partial}{\partial z} & \log P_{\phi}(y = j|c, z)\label{eq:steering}
\end{align}
Here, $\eta$ is a hyperparameter that controls the step size of the gradient update.}

\section {Experimental Setup} \label{sec:exp}
\eat{We evaluate our approach on feedback generation in educational settings, focusing specifically on cognitive complexity transformation where feedback is adjusted according to Bloom's taxonomy~\citep{blooms}. }

We evaluate our proposed steering framework on two distinct text generation domains: Textualized Gridworld, constrained path planning, and the cognitive style  (stylistic task) dataset.

\eat{Due to data scarcity, we generate a synthetic dataset using proprietary LLM ( Claude 3.5 Sonnet (v1)) through prompt engineering, a commonly adopted practice in this domain (see~\citep{liu2023pre,lu2023machine} for a comprehensive
survey). 

We also employ proprietary LLM (Claude 3.5 Sonnet (v1)) as a reference evaluator to categorize the post-steering generated feedback according to Bloom's taxonomy cognitive levels.}

\eat{The effectiveness of our steering approach is visualized through stacked bar charts, where each bar illustrates how the steered generations are distributed across different cognitive levels in the dataset.}

\subsection{Datasets}
\paragraph{Textualized Gridworld Planning.}
We utilize the Textualized Gridworld framework~\citep{kim2024language}, adapted from~\citet{zai2020deep}. To evaluate performance, we constructed the datasets with variable dimensions $M \times N$ ($M \leq 10, N \leq 10.$) for standard 
training and evaluation.\footnote{\label{fn:gridworld-extrapolate}We also evaluate on the
extrapolation set provided by \citet{kim2024language}, 
containing larger grids $M\times N$ where  either $M$ or $N \geq 10$, to test the model's ability 
to generalize steering to longer, more complex paths. Please see Table{\ref{tab:gridworld-extrapolate}} for more details} We processed the raw environments by normalizing global coordinates into a 1-based local reference frame, removing pits to focus on wall avoidance, and converting the grid state into an ASCII-based prompt. For each grid, we generated three ground-truth paths using graph search algorithms: Short, Safe, and Long. To ensure distinct planning constraints, we filtered the dataset to retain only samples where these three paths were mutually distinct with unique risk profiles. The final dataset comprises 15,250 unique examples. We split the data into training(70\%)/validation(10\%)/test (20\%). We evaluate steering performance under strict, rule-based criteria described below. 
We evaluate three grid navigation objectives: shortest, safest (minimum wall adjacency), and longest path without revisits. Optimal shortest and safest paths are computed via BFS and Dijkstra’s algorithm, and predictions are successful if they match the optimal objective value. For the NP-hard longest path task, we approximate the optimum using beam search ($B=512$) and count success if the predicted length meets or exceeds this bound. All outputs must satisfy strict path validity constraints. We report success rates over all instances.

\paragraph{Educational Cognitive Style.}
To evaluate steering on soft stylistic constraints, We construct a novel dataset to evaluate nuanced style transformation. The dataset targets ten fundamental C++ programming concepts (recursive functions, arrays, call-by-reference, switch statements, do-while loops, for-loops, while-loops, pointers, strings, and vectors), each represented by multiple coding questions and systematic feedback variations. At the initial stage, we prompt the LLM to generate 20 questions spanning easy, medium, and hard difficulties, each with its correct answer. Then, we have the LLM create four variations of each answer that contain deliberate errors. Finally, we generate feedback for both the original correct answer and all variants, using Bloom's taxonomy levels :Remember, Understand, Apply, Analyze, Evaluate, and Create~\citep{blooms}. \eat{For detailed strategy, please refer to section \ref{sec: datacretae} }


\par

\begin{table}[t]
\caption{Planning performance on Textualized Gridworld under hard constraints. Each cell reports results for Short/Safe/Long targets. Success rates are higher better, while logic violation rates are lower better.}
\label{tab:gridworld-main}
\begin{center}
\begin{sc}
\footnotesize

\begin{tabular}{@{}lccc ccc@{}}
\toprule
& \multicolumn{3}{c }{Success rates (\%) $\uparrow$}
& \multicolumn{3}{c}{Logic violation rates (\%) $\downarrow$} \\
\cmidrule(lr){2-4} \cmidrule(lr){5-7}
Method & Qwen & Phi & Llama & Qwen & Phi & Llama \\
\midrule
ICL
& 29.3/3.7/3.2
& 12.1/3.3/2.9
& 26.9/2.7/3.2
& 57.1/55.5/76.0
& 81.3/76.0/89.1
& 52.1/51.2/62.2 \\

CAA
& 53.9/4.2/5.6
& 38.6/8.7/4.9
& 39.8/4.2/4.1
& 38.8/39.6/19.9
& \textbf{36.4}/\textbf{38.9}/\textbf{38.2}
& \textbf{38.8}/\textbf{39.6}/\textbf{33.9} \\

ITI
& 32.9/9.2/6.5
& 38.8/4.7/3.3
& 39.2/6.4/4.0
& 57.8/\textbf{18.3}/19.2
& 39.9/48.5/40.3
& 67.5/65.9/43.7 \\

SAE-SSV
& 54.8/14.2/8.2
& 35.8/9.6/4.9
& 39.5/7.9/4.8
& 24.7/23.4/24.4
& 55.0/55.8/55.8
& 46.5/47.5/48.8 \\

\midrule
Dense-opt
& 54.0/14.8/7.8
& 38.0/9.4/4.9
& 39.9/9.3/5.0
& 26.7/23.5/24.7
& 54.7/55.0/55.6
& 48.0/45.8/48.8 \\

SAE-opt-anch
& \textbf{56.0}/14.5/7.4
& 37.4/9.1/4.0
& 39.5/7.1/4.9
& 23.4/27.0/\textbf{18.9}
& 56.1/57.0/55.5
& 47.7/47.4/47.0 \\

SAE-opt
& 55.5/\textbf{15.2}/\textbf{8.4}
& \textbf{38.9}/\textbf{10.4}/\textbf{5.2}
& \textbf{40.4}/\textbf{9.6}/\textbf{5.9}
& \textbf{21.0}/20.1/21.6
& 54.2/52.3/52.2
& 43.0/43.0/42.2 \\
\bottomrule
\end{tabular}%
\end{sc}
\end{center}
\vskip -0.15in
\end{table}

Our dataset comprises 3,000 data points, calculated as follows: 10 topics × 20 questions × 3 difficulty levels × 5 code variants (one correct, four with errors). We split the data by topic into training (50\%), validation (20\%), and test (30\%) sets to ensure conceptual separation between splits. 

\paragraph{TruthfulQA.}
To complement our two primary domains with an established external benchmark, we evaluate on
TruthfulQA~\citep{lin2022truthfulqa} using its open-ended generation setup. 

\eat{
\subsection{Variations in verbosity}
To create length-varied feedback, we employed a three-stage prompting pipeline using LLMs: 
\begin{inparaenum}[(1)]
\item \textbf{Base Questions:} LLM, acting as a C++ expert, generated questions spanning ten function types and three difficulty levels (easy, medium, hard).
\item \textbf{Code Variations:} LLM, assuming the role of a student, produced four error-prone code variations for each question.
\item \textbf{Feedback Generation:} LLM, as a code review assistant, provided feedback at three length granularities (short,medium, long) for both the original and the modified codes.
\end{inparaenum}

\subsection{Feedback Cognitive Variations}
For cognitive-style feedback, we guided an LLM configured as a computer science instructor to produce responses aligned with Bloom’s Taxonomy~\citep{uugur2015self}. Bloom's Taxonomy classifies educational learning objectives into six hierarchical levels: Remembering, Understanding, Applying, Analyzing, Evaluating, and Creating, progressing from lower-order thinking skills to higher-order thinking skills, reflecting a range of pedagogical depth.

For all future reference, we will refer to these two datasets as  "style" data.
}

\subsection{Models and Configurations} 
We evaluate constrained path planning on three publicly released open-weight LLMs: Qwen3-4B-Instruct (Qwen), Phi-3-Mini-4K-Instruct (Phi), and LLaMA3.1-8B-Instruct (LLaMA;\citealp{qwen3technicalreport,abdin2024phi,llama3modelcard}). For the Qwen and Phi models, we use the standard Alpaca~\citep{alpaca} prompting format, with a maximum context size of 512 tokens. For LLaMA, we use its corresponding instruction format while keeping the same context length. For each task, we perform one epoch of supervised fine-tuning on the training split to ensure task competence (\ie correct path syntax for TGW and domain-appropriate feedback for cognitive style adaptation). This fine-tuning stage is performed once per model and task. After fine-tuning, all base model parameters are frozen for the remainder of the experimental pipeline. For cognitive style adaptation we primarily use the Phi model~\citep{abdin2024phi} as the base model.\par 

SAEs are trained offline on internal activations of the frozen, fine-tuned base models. We extract query activations from a single layer (middle layer \ie for Qwen it is layer 17 and for Phi and LLaMA it is 15) and treat each attention head independently, training one SAE per query attention head. For Phi, each query head has dimensionality 96 and is mapped to a latent space of dimension 256. For Qwen and LLaMA, each query head has dimensionality 128 and is mapped to a latent space of dimension 512.\par

SAEs are trained using the objective defined in equation \ref{eq:sae_loss}, which combines reconstruction loss, $L_{1}$ sparsity on latent activations, and bias decay. We evaluate $L_{1}$ regularization coefficients $
\lambda \in \{ 3 \times 10^{-3},\; 3 \times 10^{-2},\; 3 \times 10^{-1} \}$.  Training is performed for 40 epochs using the Adam Optimizer. We employ a linear warm-up followed by a cosine decay scheduler to dynamically adjust the learning rate, starting from \(3 \times 10^{-5}\). Optimization is performed using the Adam optimizer with \(\beta_{1} = 0.90\) and \(\beta_{2} = 0.99\). Unless otherwise stated, all reported results use
$\lambda$=0.003.

During inference-time steering, we use $\eta=$ 0.5 for LLaMA and Qwen and $\eta=$ 0.8 for Phi for the TGW task. For the cognitive style dataset we use $\eta=$ 0.8. For convergence we consider $\epsilon = 10^{-3}$.
\eat{
We study two experimental setups for steering towards a desired style: \textit{(i)} steering from a source style and \textit{(ii)} steering from LLM output that may belong to any kind of style.}

\subsection{Baselines}




We refer to our proposed method as \textbf{SAE-OPT}. We also evaluate two variants to isolate the effects of anchoring and sparsity. The first variant, \textbf{SAE-OPT-ANCH}, augments the prototype-softmax steering objective with an $L_2$ anchor that penalizes deviation from the original latent encoding. Let
\[
z_0 = \mathrm{Enc}\!\left(\mathrm{LLM}_{\leq \ell}(x,y)\right)
\]
denote the unsteered sparse latent representation at intervention layer $\ell$, and let $k_T$ denote the target attribute. The anchored objective is
\[
z^*=\arg\max_{z}\; \log P_\phi(a=k_T \mid z)\;-\;\gamma\|z-z_0\|_2^2,
\]
where $\gamma \ge 0$ controls the trade-off between steering strength and preservation of the original encoding. Because $P_\phi$ is defined through a softmax over distances to all class prototypes, this objective still accounts for the relative geometry of all attribute classes; the anchor only constrains the magnitude of the latent update.

The second variant, \textbf{DENSE-OPT}, applies the same gradient-based steering objective directly to the dense query activation at intervention layer $\ell$, rather than to the sparse representation produced by the SAE. 

We also used the following baselines:\\
\textbf{ICL}~\citep{brown2020language}: For few-shot in-context learning, we include N examples sampled from the training data in our prompts and apply a few-shot learning approach~\citep{brown2020language}. captures the intended transformation. \eat{Due to constraints in context length, we prefer to use \emph{phi-3-mini-128K instruct} model for this experiment.~\footnote{We have not found any claim that confirms \emph{phi-3-mini-128K instruct} has a better performance compared to its 4K counterpart, except increased input context window.}}

\textbf{CAA}~\citep{rimsky2024steering}: Computes steering vectors as the average difference in residual stream activations between positive and negative example pairs for a behavior, then adds these vectors (with a tunable coefficient) to activations after the prompt during inference.

\textbf{Inference Time Intervention} (ITI;\citealp{li2023inference}): Identifies directions in attention heads that correspond to the target behavior (\ie truthfulness), then adds scaled vectors along these directions to the top-K most relevant heads during inference, enabling targeted steering with minimal disruption to the model.

\textbf{SAE-SSV}~\citep{he2025sae}: We adapt the steering logic from SAE-SSV by applying it to the attention query space instead of the residual stream. We extract sparse latent query features from our trained SAEs and define a static steering direction using the mean difference between target class centroids. This baseline serves as a linear control to evaluate the necessity of our proposed gradient-based optimization for handling hard constraints.

\textbf{DISCO-Q}~\citep{torop2025disco}: A static query-level steering method that adds a fixed steering vector to dense query activations. We include it as a query-site static baseline, isolating the effect of our gradient-based optimization against a static intervention at the same (query) site.


    

\subsection{ Results and discussion}

We organize the results around three sections. Section ~\ref{subsec:hardconstraint} evaluates whether sparse query-level steering satisfies hard, verifiable constraints using TGW, where both path validity and attribute optimality are rule-checkable. 
Sections~\ref{subsec:cognitivestyle} and ~\ref{subsec:axbench1} examine transfer to soft semantic control by evaluating Bloom's Taxonomy steering, first through a classifier-based analysis and then using the standardized AxBench three-axis ~\citep{wu2025axbench}. Section~\ref{subsec:gbvsas} investigates an orthogonal design question to the site selection analysis in Section~\ref{subsec:qli}. Specifically, we test whether gradient based optimization outperforms static vector addition on a shared  (TruthfulQA;\citealp{lin2022truthfulqa}), extending the accuracy distortion analysis of Figure~\ref{fig:pareto_main} from the intervention site axis to the optimization axis. Finally, Section~\ref{subsec:entanglement} examines the role of sparsity by comparing L1 and L2 autoencoder regularization, showing that L1 regularization reduces non-target prototype drift and yields cleaner steering.

\subsubsection{Hard constraint steering in Textualized Gridworld}
\label{subsec:hardconstraint}
Hard-constraint steering provides a stringent test of a steering method because success requires both target adherence and structural validity. A generated path is counted as successful only if it satisfies the intended attribute and remains valid; for example, a shorter path that enters a wall, or a safe path that is disconnected, is still a failure. Textualized Gridworld makes both conditions exactly checkable, allowing us to measure success rate and logic violations separately and interpret them jointly (Table~\ref{tab:gridworld-main}). Under this criterion, SAE-OPT achieves the highest or tied-highest success rate in most settings while maintaining among the lowest violation rates. This suggests that its gains in attribute control do not primarily come from sacrificing path validity.

Short paths align closely with the model's post-fine-tuning bias, yielding relatively high success across all methods. Safe and Long targets are considerably more difficult. On Safe, our approach improves success from 3.7\% (ICL) and 4.2\% (CAA) to 15.2\% on Qwen, with similar relative gains on Phi and LLaMA. On the NP hard Long objective, absolute performance remains modest, but our method achieves the highest success across all methods.

An observed pattern is the relative stability of violation rates under our method within each model across targets (approximately 21\% on Qwen, 52\% on Phi, and 43\% on LLaMA). This behavior is consistent with the localized nature of query-level interventions. Query-level modifications influence attention weights rather than directly altering the residual stream. As a result, steering tends to reroute paths within the valid-path manifold instead of violating structural constraints.

Methods can be grouped into distinct categories. Few-shot prompting (ICL) performs the weakest overall. Dense residual stream methods are competitive on short but degrade on safe and long. Sparse query methods perform better on the harder targets, suggesting that the sparse query space is an effective control surface. Gradient based prototype optimization generally outperforms static steering vectors on safe and long.

The relative ranking of methods is consistent across architectures and extends to larger grids in the extrapolation setting (Table~\ref{tab:gridworld-extrapolate}). Appendix~\ref{sec:app} provides quantitative visualizations through heatmaps (Section~\ref{subsec:heatmap}) and cell-level attention analysis (Section~\ref{sec:attention_analysis}), illustrating target-aligned shifts that correspond to both improved success and preserved validity.

\subsubsection{Cognitive style steering (Bloom’s Taxonomy)}

\label{subsec:cognitivestyle}

We evaluate steering over six Bloom's taxonomy levels: Remember (Rem), Understand (Und), Apply (App), Analyze (Ana), Evaluate (Eva), and Create (Cre),
using Phi as the base model. Each steered output is assessed by GPT-4o and
Claude on two dimensions: (a)~Bloom-level classification identifying the
dominant cognitive level, and (b)~binary relevance judgment of whether the
feedback engages with the input code. Table~\ref{tab:diagonal} reports diagonal hit rates from confusion
matrices under both judges. The base model exhibits a strong default bias
toward Evaluate, making bias conflicting levels particularly remember and
create the most discriminative targets. SAE-OPT achieves the highest
average hit rate under both judges, with the largest gains on precisely these
levels. The SAE-OPT-ANCH variant collapses almost entirely to Evaluate, confirming
that the L2 anchor prevents the optimizer from reaching distant prototype
regions and that the unconstrained objective is necessary for balanced steering.

Table~\ref{tab:relevance} reports per-target relevance rates. Our method
maintains the highest relevance across nearly all targets, while
residual stream methods show lower relevance,
suggesting that query level interventions redirect cognitive style without
displacing the model's grounding in the input code. We further observe that
enforcing sparsity reduces blending of multiple cognitive levels within a
single response, improving single-level alignment over dense query steering.
Full confusion matrices, relevance rates, and qualitative comparisons appear in
Appendix~\ref{app:confusion} (see Section~\ref{sec:Qgeneration},~\ref{sec:steered_ft} for qualitative examples)

\begin{table}[t]
\centering
\caption{Diagonal hit rates (\%) from output-class confusion matrices under GPT-4o / Claude judges. 
Each cell shows the percentage of steered generations correctly classified as the intended target. 
Full confusion matrices are in Appendix~\ref{app:confusion}.}
\label{tab:diagonal}
\small
\begin{sc}
\footnotesize
\setlength{\tabcolsep}{4pt}
\begin{tabular}{@{}ll cccccc c@{}}
\toprule
\textbf{Method} & \textbf{Judge} & \textbf{Rem.} & \textbf{Und.} & \textbf{App.} & \textbf{Ana.} & \textbf{Eva.} & \textbf{Cre.} & \textbf{Avg.} \\
\midrule
\multirow{2}{*}{CAA} 
  & GPT-4o & 9.7  & 21.5 & 1.6  & 46.4 & 45.6 & 8.3  & 22.2 \\
  & Claude & 13.1 & 15.1 & 1.1  & 55.6 & 46.2 & 19.1 & 25.0 \\
\midrule
\multirow{2}{*}{ITI} 
  & GPT-4o & 7.6  & 24.7 & 4.8  & 32.0 & 17.5 & 9.5  & 16.0 \\
  & Claude & 5.9  & 15.0 & 5.6  & 21.2 & 12.8 & 0.1  & 10.1 \\
\midrule
\multirow{2}{*}{SAE-SSV} 
  & GPT-4o & 1.4  & 21.2 & 1.9  & 19.3 & 46.7 & 12.0 & 17.1 \\
  & Claude & 2.9  & 16.9 & 3.5  & 20.0 & 42.8 & 17.8 & 17.3 \\
\midrule
\multirow{2}{*}{DiscoQ} 
  & GPT-4o & 2.5  & 18.4 & 1.4  & 19.1 & 44.0 & 13.8 & 16.5 \\
  & Claude & 3.9  & 14.0 & 0.4  & 19.8 & 43.7 & 18.3 & 16.7 \\
\midrule
\multirow{2}{*}{Dense-opt} 
  & GPT-4o & 8.5  & 25.1 & 5.2  & 20.5 & 52.5 & 14.6 & 21.1 \\
  & Claude & 12.2 & 16.3 & 7.2  & 27.1 & 48.5 & 16.5 & 21.3 \\
\midrule
\multirow{2}{*}{SAE-opt-anch} 
  & GPT-4o & 1.6  & 4.5  & 4.1  & 16.3 & \textbf{80.2} & 3.3  & 18.3 \\
  & Claude & 2.8  & 3.6  & 5.2  & 12.4 & \textbf{83.6} & 5.5  & 18.9 \\
\midrule
\multirow{2}{*}{SAE-opt} 
  & GPT-4o & \textbf{10.8} & \textbf{28.9} & \textbf{8.4} & \textbf{31.0} & 54.7 & \textbf{18.4} & \textbf{25.4} \\
  & Claude & \textbf{16.5} & \textbf{19.3} & \textbf{10.8} & \textbf{35.4} & 55.8 & \textbf{24.7} & \textbf{27.1} \\
\bottomrule
\end{tabular}
\end{sc}
\end{table}

\subsubsection{Quantitative evaluation with AxBench}
\label{subsec:axbench1}

To complement classifier based analysis with a standardized evaluation protocol, 
we report results on AxBench~\citep{wu2025axbench}, which scores steered 
outputs along three axes, concept incorporation ($s_c$), instruction 
following ($s_i$), and fluency ($s_f$),aggregated via harmonic mean 
(HM $\in [0,2]$). Methods that force a concept at the expense of 
coherence are heavily penalized, making it a stricter test than 
single-label classification.(see Appendix~\ref{subsec:axbench})

Our method achieves the highest HM on 5 of 6 Bloom levels 
(\ie Understand: 0.950 vs.\ 0.907 for DENSE-OPT and 0.752 for CAA; 
Create: 0.764 vs.\ 0.697; Evaluate: 1.399 vs.\ 1.389; full per-axis 
breakdown in Table~\ref{tab:axbench_full}, Appendix~\ref{app:axapp}). 
All methods struggle on Remember (HM $< 0.51$), which requires 
suppressing the model's default tendency to elaborate. On 
higher-complexity levels (Understand--Create), gains stem from improved 
concept adherence ($s_c$) without sacrificing instruction following 
or fluency, confirming that gradient ascent on middle-layer query 
latents provides the localized control needed for nuanced stylistic 
steering.

\subsubsection{ Gradient-based versus static vector addition steering on TruthfulQA}
\label{subsec:gbvsas}

Section~\ref{subsec:qli} (Figure~\ref{fig:pareto_main}) show that query level interventions incur lower distributional cost than residual stream interventions at matched success, establishing an accuracy distortion trade off along the intervention site axis. We also investigate whether a similar trade-off exists along the optimization axis (\ie whether gradient-based updates incur lower distortion than static vector addition). To evaluate this choice  on an established benchmark (Table ~\ref{tab:truthfulqa}), we compare against static steering baselines on TruthfulQA~\citep{lin2022truthfulqa} using the open-ended generation setup of~\citet{torop2025disco} and the \%Truthful × \%Informative (\%T×I) metric~\citep{lin2022truthfulqa}. The static baselines span different intervention: CAA uses residual stream, ITI uses attention heads, and DISCO-Q uses dense query, so any advantage of gradient-based optimization is attributable to the update rule rather than the site.

\begin{table}[h!]
\caption{Gradient-based versus static vector addition steering on TruthfulQA. We report steering accuracy (\%T$\times$I) and distributional cost (JSD) across steering methods on two base models.}
\label{tab:truthfulqa}
\centering
\small
\begin{sc}
    \footnotesize
\begin{tabular}{llcccc}
\toprule
\textbf{Method} & \textbf{Type} & \textbf{\%T$\times$I $\uparrow$} & \textbf{Mean JSD $\downarrow$} & \textbf{Median JSD $\downarrow$} & \textbf{Max JSD $\downarrow$} \\
\midrule
Qwen & -- & 0.6749 & -- & -- & -- \\
ITI & Static & 0.6778 & 0.009839 & 0.006841 & 0.082882 \\
CAA & Static & 0.6914 & 0.107478 & 0.102157 & 0.278795 \\
Disco-Q & Static & 0.7078 & 0.030830 & 0.027103 & 0.131915 \\
Dense-opt & Gradient & \textbf{0.7407} & \textbf{0.002392} & \textbf{0.000731} & \textbf{0.028518} \\
\midrule
LLaMA & -- & 0.7366 & -- & -- & -- \\
ITI & Static & 0.7423 & 0.033181 & 0.019546 & 0.198342 \\
CAA & Static & 0.7572 & 0.063668 & 0.049712 & 0.239238 \\
Disco-Q & Static & 0.7866 & 0.082272 & 0.074456 & 0.227702 \\
Dense-opt & Gradient & \textbf{0.7928} & \textbf{0.001314} & \textbf{0.000873} & \textbf{0.015991} \\
\bottomrule
\label{tab:gdvssv}
\end{tabular}
\end{sc}

\end{table}

Across both models, DENSE-OPT achieves the highest \%T×I while reducing mean JSD by over an order of magnitude compared to the static baseline. Notably, maximum JSD remains below $0.03$ for DENSE-OPT, whereas static methods exhibit peaks exceeding $0.13-0.28$, confirming that gradient based optimization avoids the worst case distributional shifts inherent to static vector addition. Table \ref{tab:gdvssv} corroborates the accuracy distortion trade off observed in Figure \ref{fig:pareto_main}, now validated on an external benchmark with independent baselines.

\subsubsection{Sparsity reduces steering entanglement}
\label{subsec:entanglement}

We test whether sparse regularization provides a measurable advantage during gradient-based steering. 
To isolate the effect of sparsity, we compare the SAE ($L_{1}$) against an otherwise identical autoencoder trained with an $L_{2}$ regularization penalty. 
This comparison asks whether $L_{2}$ latents are sufficient for clean steering, or whether $L_{1}$ regularization improves the locality of the intervention.

We measure locality using non-target drift, the absolute change, from the initial to final optimization step, in normalized distance to prototype classes that are not targeted by the optimizer.

For a subset of the test dataset, we steer the representation toward each of the three path classes at two regularization strengths, 0.003 and 0.03, where the $L_{1}$ model exhibits meaningful sparsity. 
During optimization, we record the normalized distance to all class prototypes at every step.

\begin{figure}[t]
    \centering
    \includegraphics[width=0.85\textwidth]{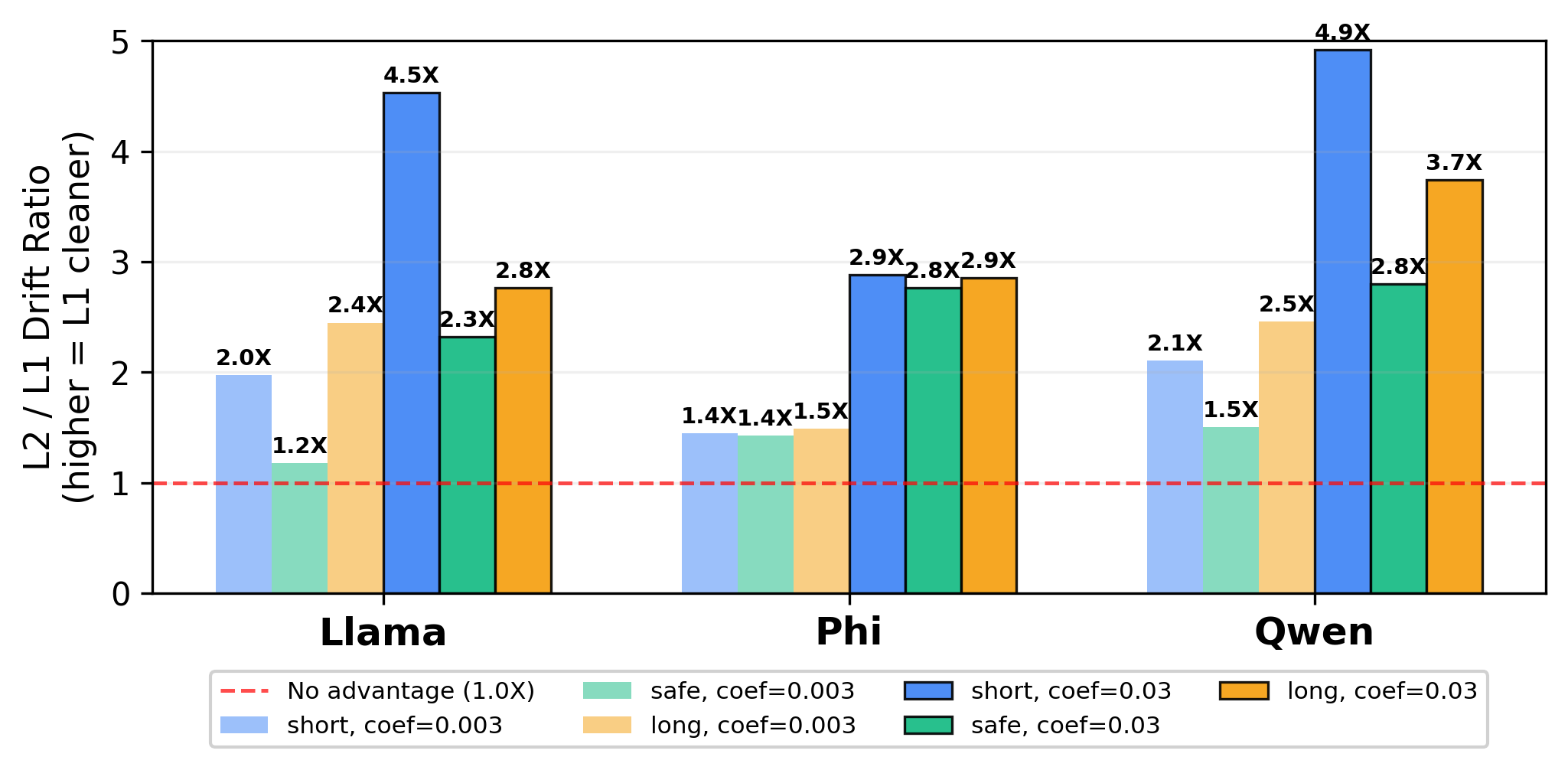}
   \caption{
L2/L1 non-target drift ratio across three models (Qwen, Phi, Llama), three target classes, and two regularization coefficients. 
Bars above the dashed red line indicate that L2 induces more non-target drift than L1, meaning L1 steers more cleanly. 
Faded bars correspond to coef$\,=\,$0.003; bold-outlined bars correspond to coef$\,=\,$0.03. 
Across configurations, $L_{1}$ consistently reduces collateral drift, with the advantage becoming strongest at the larger coefficient, where $L_{2}$ induces up to 4.9$\times$ more non-target drift.
}
    \label{fig:drift}
\end{figure}

\textbf{$L_{1}$ reduces collateral drift across models} (Figure~\ref{fig:drift}). 
Across the regularization settings, $L_{1}$ produces lower non-target drift. 
This effect appears across all three models and all target classes, suggesting that the sparsity advantage is not specific to a single architecture or target behavior. 
 
Thus, the advantage of $L_{1}$ is not that it can steer while $L_{2}$ cannot; rather, $L_{1}$ provides a cleaner steering interface by achieving comparable target convergence with less collateral movement in non-target prototype geometry.

\section{Related works}
\subsection{Sparse Autoencoders}
\label{sec:sae_bg}
Sparse coding, introduced by Mallat et al.~\citep{mallat1993matching} and advanced by Olshausen et al.~\citep{olshausen1996emergence} for unsupervised learning, was initially applied to image processing~\citep{liu2015sparse}. Parallelly, Hinton et al.~\citep{hinton2006reducing} introduced autoencoders for dimensionality reduction, leading to the convergence in sparse autoencoders (SAEs)~\citep{konda2014zero,lee2007sparse} that enforce sparsity via $L_1$ regularization to extract meaningful features.SAEs have gained traction for decomposing LLM activations into interpretable features~\citep{cunningham2023sparse}, representing semantic and syntactic concepts as linear combinations in models like BERT~\citep{yun2021transformer}, GPT~\citep{sharkey2022taking}, and Pythia~\citep{cunningham2023sparse}~\citep{bloom2024open,marks2023some}. This aligns with the Linear Representation Hypothesis~\citep{mikolov2013linguistic,nanda2023progress,park2023linear}, favoring monosemantic neurons~\citep{bau2020understanding}. However, networks encode more features than neurons, resulting in polysemanticity~\citep{scherlis2022polysemanticity,mu2020compositional} explained by the superposition hypothesis~\citep{elhage2022toy}. Recent advancements scale SAEs for better monosemanticity and disentanglement, aiding interventions in sparse spaces~\citep{bayat2025steering,he2025saif}. Methods such as activation pruning~\citep{rajamanoharan2024jumping} or thresholds~\citep{rajamanoharan2024improving} balance sparsity and reconstruction quality.\par

 Building on these foundations,~\citet{marks2025sparse} demonstrate that SAE features can be organized into sparse feature circuits, causally implicated subnetworks of interpretable features enable both mechanistic understanding and targeted behavioral editing. This circuit level perspective complements further extensions of SAEs to steering include vector refinement via semantic denoising~\citep{wang2025enhancing},correlated feature steering at generation time~\citep{cho2025corrsteer} and step level context-aware SAEs for reasoning~\citep{yang2026step}. ~\citet{soo2025interpretable}propose Feature Guided Activation Additions (FGAA), which combines contrastive activation addition with SAE targeted steering by optimizing over SAE features to produce interpretable yet precise steering vectors.

\par
\subsection{Steered Generation}
Content generation is controllable through latent space~\citep{larsen2016autoencoding,white2016sampling} and embedding adjustments~\citep{han2024word}. However, prevalent methods such as fine-tuning and RLHF~\citep{ranzato2015sequence,dathathri2019plug,ilharco2022editing,meng2022locating} modify model weights, which risks degrading performance~\citep{qi2023fine}. Alternatives avoiding weight changes intervene at decoding~\citep{gu2017trainable}, prompt~\citep{zhou2022steering}, or token embedding levels~\citep{khashabi2021prompt}.\par

\textbf{Activation interventions} offer efficient control without weight modifications, using steering vectors~\citep{hernandez2023inspecting,subramani2022extracting} and latent arithmetic~\citep{correia2019adaptively}. Inference-Time Intervention (ITI)~\citep{li2024inference} promotes truthfulness by adding probe-derived vectors to attention heads, whereas ActAdd~\citep{turner2023activation} steers behavior using activation differences between contrasting prompts, even application in personalized text generation~\citep{zhang-etal-2025-personalized}. Similarly, ~\citet{liu2023context} apply analogous vector interventions to enhance in context learning.  Beyond static single-vector addition, richer parameterizations of the intervention itself have recently emerged. LinEAS~\citep{rodriguez2025end-to-end} learns end-to-end activation steering maps via a distributional loss that jointly accounts for layer wise shifts, improving robustness. Further methods show that combining nonlinear predictors across multiple layers yields universal steering and monitoring that generalizes across concepts~\citep{beaglehole2026toward}. On the optimization axis, one-shot gradient-based optimization of steering vectors has been shown to mediate safety-relevant behaviors with broad transfer~\citep{dunefsky2025oneshot}.

\textbf{Attention interventions} Residual stream interventions can broadly perturb activations and overwrite context. Consequently, a growing body of work instead intervenes at the attention mechanism itself. The PASTA method~\citep{zhang2024tell} profiles attention heads that benefit task performance and statically reweights attention within those specific heads
, whereas SpotLight~\citep{Venkateswaran2025SpotlightYI} instead dynamically amplifies attention scores on instruction tokens across all layers and heads to ensure consistent attention allocation. In a parallel work, JoLA learns sparse, gated head edits with mixed edit forms~\citep{lai2025joint}, while ~\citet{Nguyen2025MultiAttributeSO} introduces attribute-specific steering vectors with gating mechanisms for multi-attribute control. Further refining head selection, ~\citet{zhan2025deal} proposes a causal-attribution framework that disentangles behavior relevant head activations and computes head-wise confidence scores for principled module selection. At the reasoning level, CREST identifies cognitive attention heads linked to non-linear reasoning patterns and applies test-time interventions to steer reasoning trajectories~\citep{Zhang2025UnderstandingAS}. Concurrent with our work, similar benefits from query and attention-level interventions have been reported ~\citep{torop2025disco}. ~\citet{davarmanesh2026efficient} introduce attention guided feature learning that automates token selection, accounts for feature heterogeneity.

\textbf{SAE based steering} Despite SAE training costs~\citep{gao2024scaling}, their interpretability supports feature-based steering~\citep{durmusevaluating},  editing middle-layer residual streams to resolve knowledge conflicts~\citep{zhao2024steering} via sparsity~\citep{chalnev2024improving,o2024steering}. Recent extensions include supervised steering in sparse subspaces with classifiers for sentiment/truthfulness~\citep{he-etal-2025-sae}.SAE-based denoising of concept vectors for more reliable control~\citep{zhao-etal-2026-denoising}, and prototype-aligned sparse steering for vision, which conceptually parallels our prototype-based formulation but operates in a different modality~\citep{Chatzoudis2025VisualSS}.Recent studies apply SAEs to reasoning benchmarks, with specific instances like amplifying targeted reasoning features to improve benchmark performance~\citep{galichin2025have} and utilizing token  wise decaying steering strategies to reliably elicit step by step mathematical reasoning~\citep{xie2025comparative}.
Other interpretability frameworks focus on unsupervised multi-concept shifts~\citep{joshi2025identifiable} or concept bottlenecks~\citep{kulkarni2025interpretable}. In the safety and control domain, approaches prioritize mutual information based explanations for anti-jailbreaks~\citep{wu2025interpreting}.  Separately, in the multilingual domain, researchers use directional feature ablation to uncover language specific capabilities~\citep{deng2025unveiling} and single feature steering to causally control the target generation language~\citep{chou2025causal}. Furthermore, efficiency-centric methods utilize feature-level constrained DPO~\citep{Yin2024DirectPO} or hypothesis generation from embeddings~\citep{movvasparse}, while RL-specific works reveal TD errors in residual streams~\citep{demircansparse} or analyze CoT-induced attention sparsity~\citep{wen2024sparse}.Recent attempts also  target cross-model concept alignment via universal SAEs~\citep{thasarathan2025universal} for visual tasks. Complementary prototype based approaches have also emerged~\citep{kayan2025prototype}. Concurrent work has further explored SAE-based reasoning control~\citep{fang2026controllable} and  efficient feature selection for steering~\citep{arad2025saes}, reinforcing the value of sparse level interventions.

Building on these lines of work, we investigate sparse, query level interventions for controllable generation. We train independent SAEs on attention query activations at middle layers, where relational and structural information is most accessible~\citep{vig2019analyzing}. During inference time, we perform  gradient ascent to align sparse query features with class prototype distributions. This combines query level locality, sparse disentanglement, and gradient-based adaptivity in a single framework.

\section{Conclusion and Future Work}
We propose a sparse feature based  steering method that performs inference-time control by optimizing sparse query representations in attention layers. The approach achieves reliable alignment with target attributes while preserving structural validity, outperforming residual-stream and static steering baselines under hard planning constraints. On Textualized Gridworld, it improves success rates , and it generalizes to soft semantic control by steering cognitive complexity in educational feedback. Across tasks, middle-layer query features emerge as the most effective control surface, supporting sparse query-level optimization as a principled mechanism for controllable generation.

\bibliography{tmlr}
\bibliographystyle{tmlr}

\newpage

\appendix
\onecolumn

\section{Appedix}

\label{sec:app}


\paragraph{Organization of the Appendix.}
Table~\ref{tab:appendix-guide} summarizes the contents of each appendix section for quick reference.
 
\begin{table}[h!]
\centering
\caption{Guide to the Appendix.}
\label{tab:appendix-guide}
\small
\renewcommand{\arraystretch}{1.35}
\begin{tabular}{|l|l|l|p{6.2cm}|}
\hline
\textbf{Section} & \textbf{Domain} & \textbf{Type} & \textbf{Contents} \\
\hline\hline
A.1 & TGW & Visualization & Attention heatmaps evolving across gradient-ascent steps for Short, Safe, and Long targets. \\
\hline
A.2 & TGW & Quantitative & Cell-level attention analysis with numerical attention weights under each steering target. \\
\hline
A.3 & TGW & Evaluation & Extrapolation results on larger grids ($M$ or $N \geq 10$) using Qwen3-4B. \\
\hline
A.4 & TGW / Bloom & Ablation & Direct target-center assignment vs.\ gradient-based optimization. \\
\hline\hline
B & Bloom & Qualitative & Steered generation examples (Apply, Create, Remember) comparing base vs.\ steered outputs on Phi. \\
\hline
C & Bloom & Qualitative & Side-by-side comparisons of dense query steering (w/o SAE) vs.\ SAE-based steering with LLM judgments. \\
\hline
D & Bloom & Baseline & Effect of increasing in-context examples on cognitive-style accuracy for the base model. \\
\hline\hline
E & Bloom & Protocol & AxBench-adapted three-axis LLM judge setup: Bloom level definitions and verbatim judge prompts . \\
\hline
F & Bloom & Evaluation & Full AxBench per-axis scores ($s_c$, $s_i$, $s_f$, HM) for all methods and targets. \\
\hline\hline
G & Bloom & Evaluation & Output-class confusion matrices and per-target relevance rates  under GPT-4o and Claude judges. \\
\hline\hline
H & TGW / Bloom & Ablation & Steering effectiveness across different attention layers. \\
\hline
I & TGW & Analysis & Detailed L1 vs.\ L2 non-target drift values and optimization steps per model. \\
\hline
\end{tabular}
\end{table}
\newpage

\subsection{Attention Heatmaps for Spatial Reasoning in Textualized Gridworld}
\label{subsec:heatmap}

We visualize how query-level sparse steering progressively refines the model's attention patterns over the grid during inference-time optimization. Each figure contains four panels: the unsteered output (top-left), which is typically invalid, and three subsequent panels showing intermediate steps of gradient ascent on the sparse query latents. Cell-level attention heatmaps are computed by averaging attention weights across heads, extracting the attention distribution of the final token position, aggregating over the token span corresponding to each grid cell, and normalizing across cells to obtain the distribution $p(c)$.

\begin{figure*}[!ht]
  \centering
  \includegraphics[width=0.85\textwidth, height=0.42\textheight, keepaspectratio]{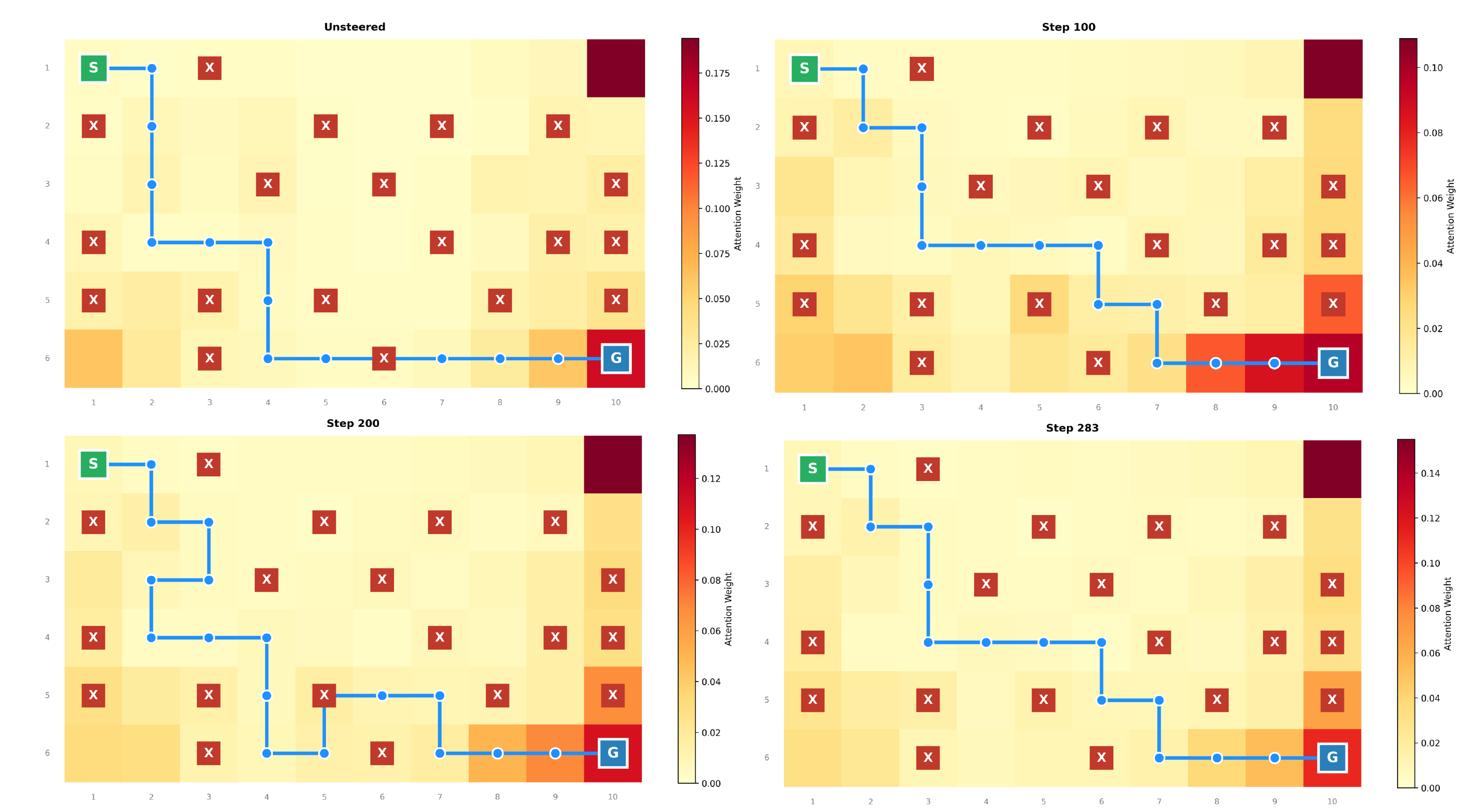}
  \caption{Evolution of attention heatmaps when steering toward the \textbf{Short} path target.}
  \label{fig:heatmap-short}
\end{figure*}

\begin{figure*}[!ht]
  \centering
  \includegraphics[width=0.85\textwidth, height=0.42\textheight, keepaspectratio]{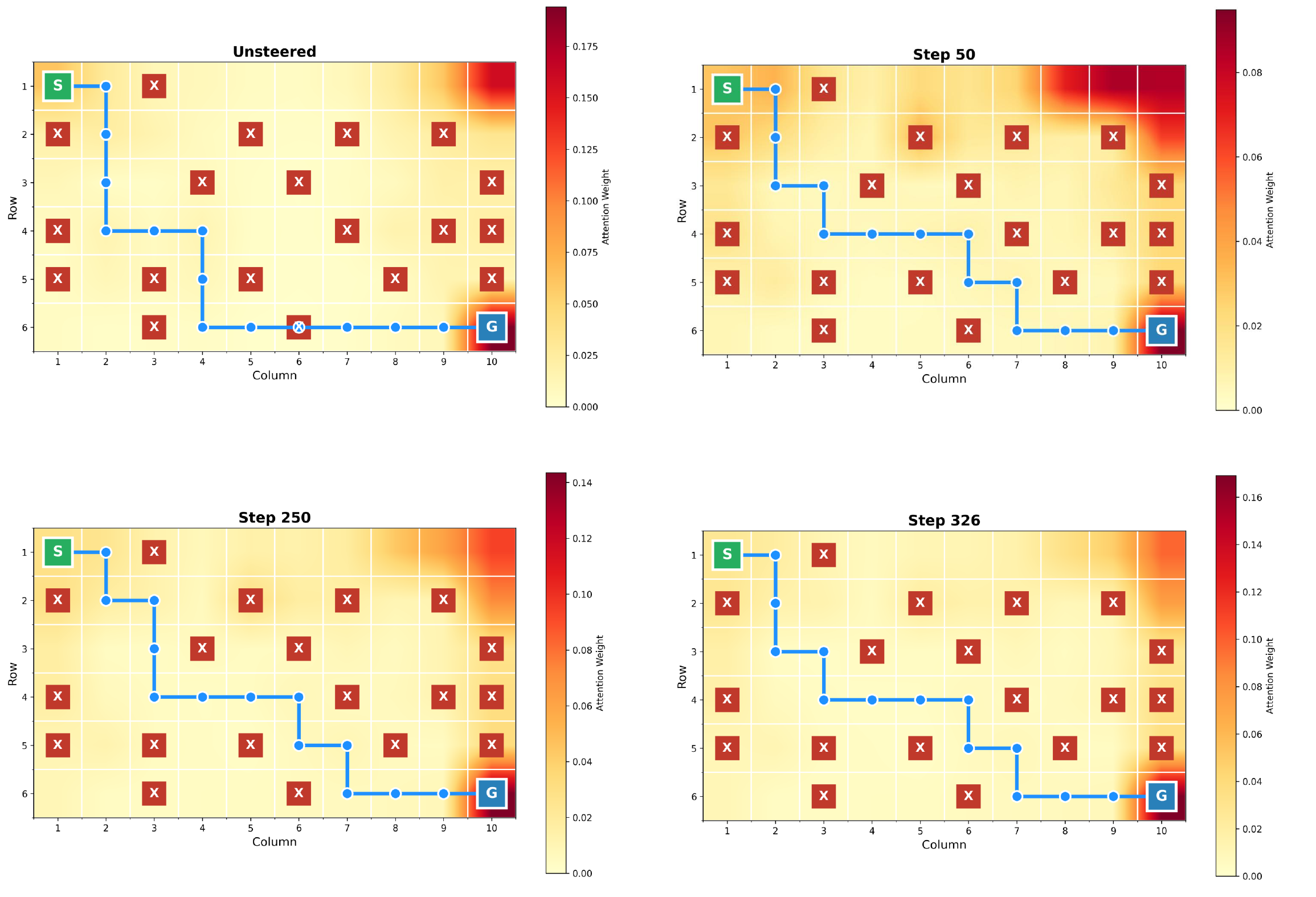}
  \caption{Evolution of attention heatmaps when steering toward the \textbf{Safe} path target.}
  \label{fig:heatmap-safe}
\end{figure*}

\begin{figure*}[!ht]
  \centering
  \includegraphics[width=0.85\textwidth, height=0.42\textheight, keepaspectratio]{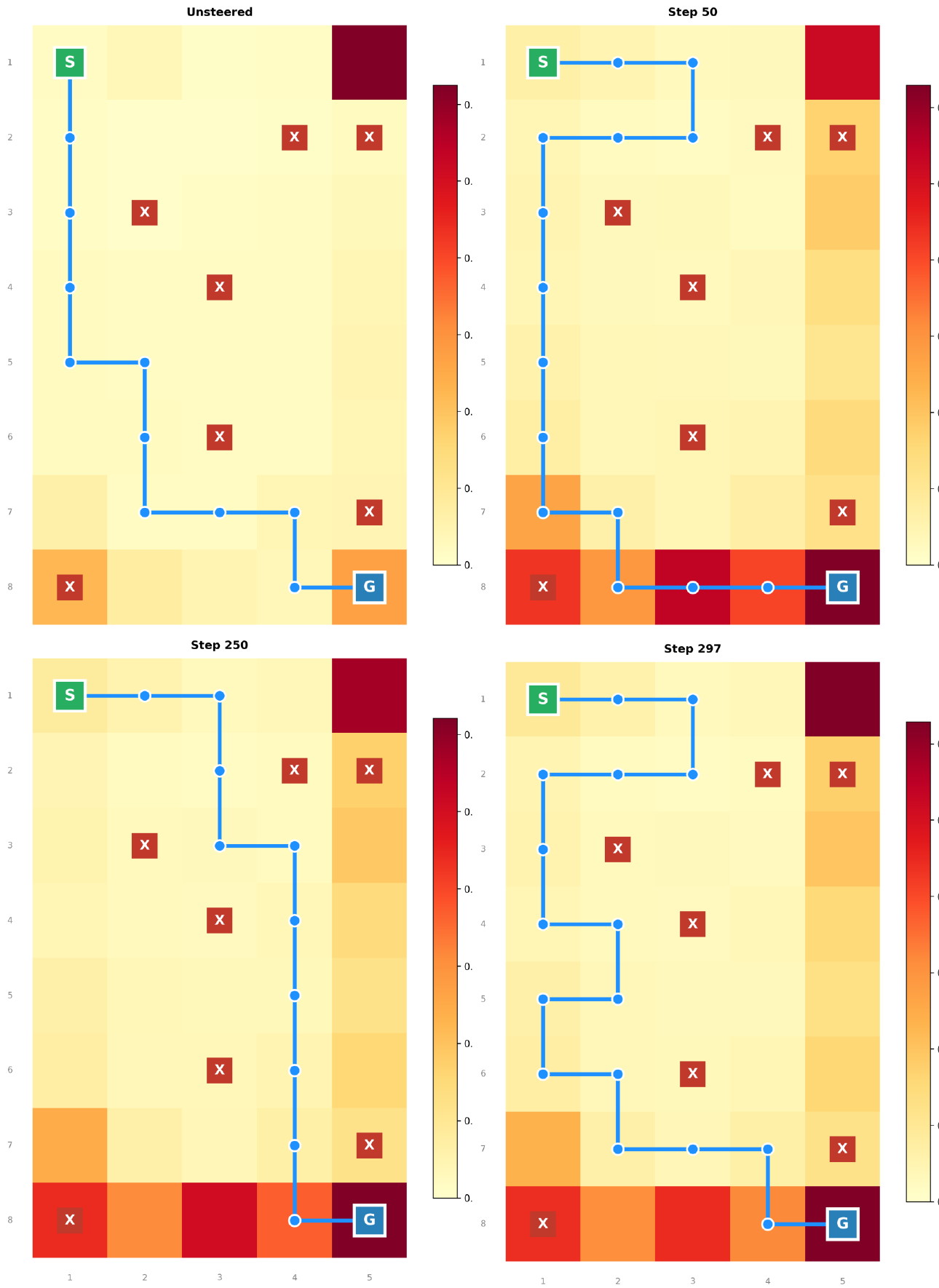}
  \caption{Evolution of attention heatmaps when steering toward the \textbf{Long} path target.}
  \label{fig:heatmap-long}
\end{figure*}
\FloatBarrier 
\subsection{Quantitative Cell-Level Attention Analysis Under Steering}
\label{sec:attention_analysis}

To understand how steering reshapes the model’s internal spatial reasoning at a mechanistic level, we examine whether the model’s attention over grid cells shifts in a manner consistent with the target attribute as gradient-based optimization progresses. For instance, when steering toward  safe paths, we expect attention to gradually stay away  from cliff cells. Similarly, short steering should concentrate attention along the most direct route, while long steering should distribute it more broadly across the grid to support extended trajectories. To test these hypotheses, we extract and visualize cell level attention distributions $  p(c)  $ at intermediate optimization steps (Figures 9-11). This analysis serves two purposes: it provides interpretable evidence that query level interventions produce spatially coherent changes in the model’s reasoning process, and it validates that steering modifies (Section~\ref{subsec:qli}) what information the model retrieves through attention.

Given the attention weight tensor \(\mathbf{A} \in \mathbb{R}^{H \times L \times L}\) from the intervention layer \(\ell\) (with \(H\) heads and sequence length \(L\)), we compute cell-level attention as follows:

\paragraph{Head Averaging.} We first average attention across all heads to obtain a more stable signal:
\begin{equation}
    \bar{\mathbf{A}} = \frac{1}{H} \sum_{h=1}^{H} \mathbf{A}_h \in \mathbb{R}^{L \times L}
\end{equation}

\paragraph{Query Selection.} We extract attention from the final generated token position:
\begin{equation}
    \bar{\mathbf{a}} = \bar{\mathbf{A}}[L-1, :] \in \mathbb{R}^{L}
\end{equation}

\paragraph{Cell Aggregation.} For each grid cell \(c\) with corresponding token indices \(\mathcal{T}_c \subseteq \{1, \ldots, L\}\), we compute the mean attention:
\begin{equation}
    \text{attn}(c) = \frac{1}{|\mathcal{T}_c|} \sum_{t \in \mathcal{T}_c} \bar{\mathbf{a}}[t]
\end{equation}

\paragraph{Normalization.} Finally, we normalize across all grid cells to obtain a probability distribution over spatial locations:
\begin{equation}
\label{eq:nAw}
    p(c) = \frac{\text{attn}(c)}{\sum_{c' \in \mathcal{C}} \text{attn}(c')}
\end{equation}
where \(\mathcal{C}\) denotes the set of all grid cells.

\textbf{How to read Figure 9-11:} Each figure contains four panels showing cell-level attention at progressive optimization steps (leftmost = unsteered baseline, rightmost = final). The number in each cell is its normalized attention weight(Equation ~\ref{eq:nAw}). X = wall (red/yellow border), S = start, G = goal, blue arrows = generated path. To track the effect of steering, compare attention values on the same cell across panels ,increasing values indicate the model is allocating more retrieval weight to that location, while decreasing values indicate reduced focus.
\newpage
\begin{figure*}[h!]
  \centering
  \label{fig:safesteer}
  
  \includegraphics[width=1\textwidth]{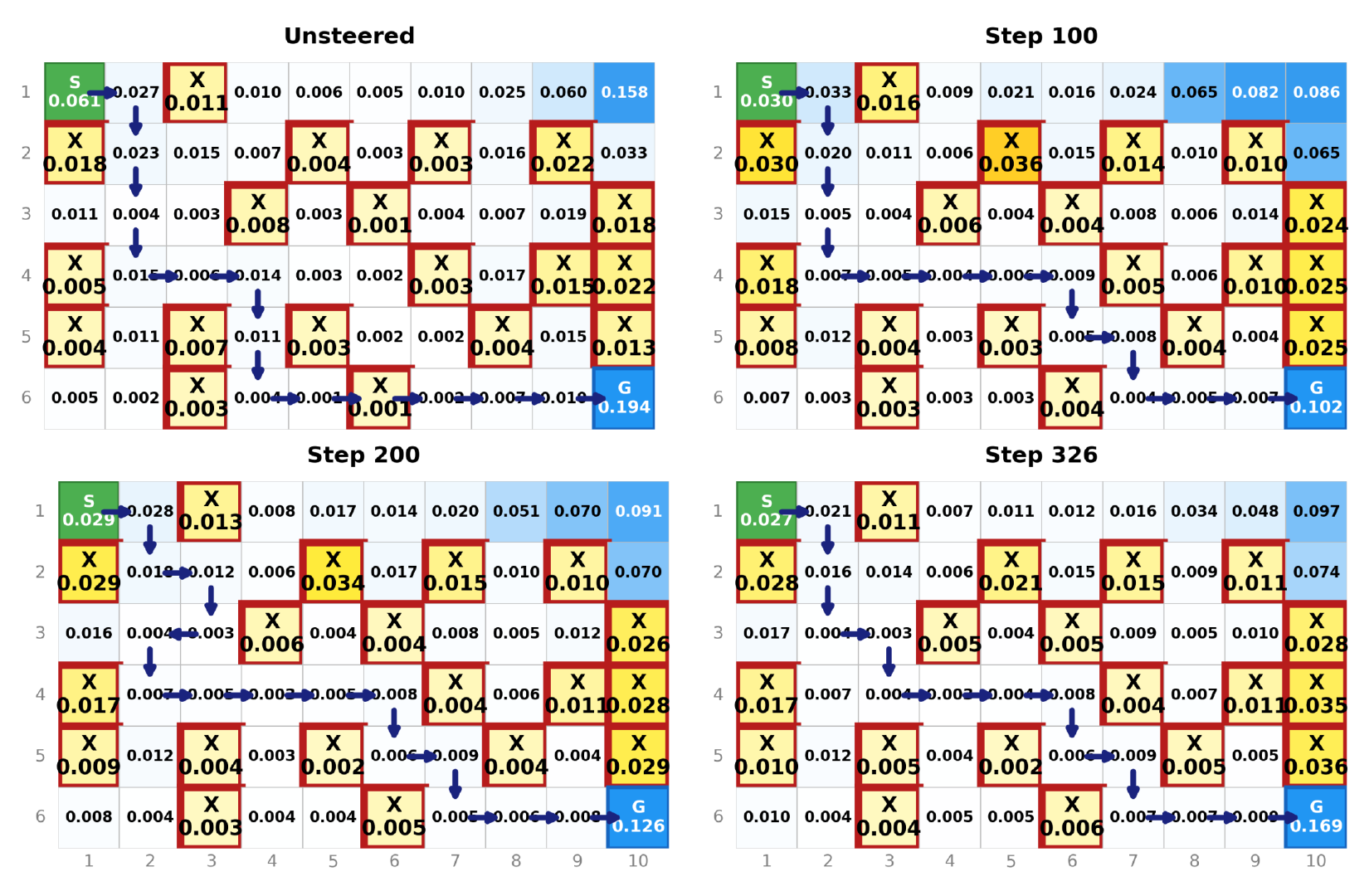}
  \caption{Cell level attention distributions during steering toward the Safe path. }
\end{figure*}

In the unsteered baseline, attention is concentrated at the endpoints (S = 0.061, G = 0.194, row 0 column 9 = 0.158), with walls and interior cells receiving minimal weight. In the unsteered phase, the model plans a direct start to goal route that passes through wall adjacent regions. Under safe steering, this endpoint dominance diminishes (S drops to 0.027, G to 0.169, row 0 column 9 to 0.097), and rest of the attention redistributes selectively. Walls near the planned safe route gain attention ( X at (1,4): 0.004 to 0.021, X at (3,0): 0.005 to 0.017, X at (4,9): 0.013 to 0.036), while interior walls far from the route stay flat or decrease (\ie, X at (2,3): 0.008 to 0.005, X at (4,4): 0.003 to 0.002). We hypothesize increased attention on a wall cell does not mean the model is drawn toward it, instead, the query mechanism retrieves obstacle location information, which then informs next-token prediction to select cells with greater clearance.
\newpage
\begin{figure*}[h!]
  \centering
  \includegraphics[width=\textwidth]{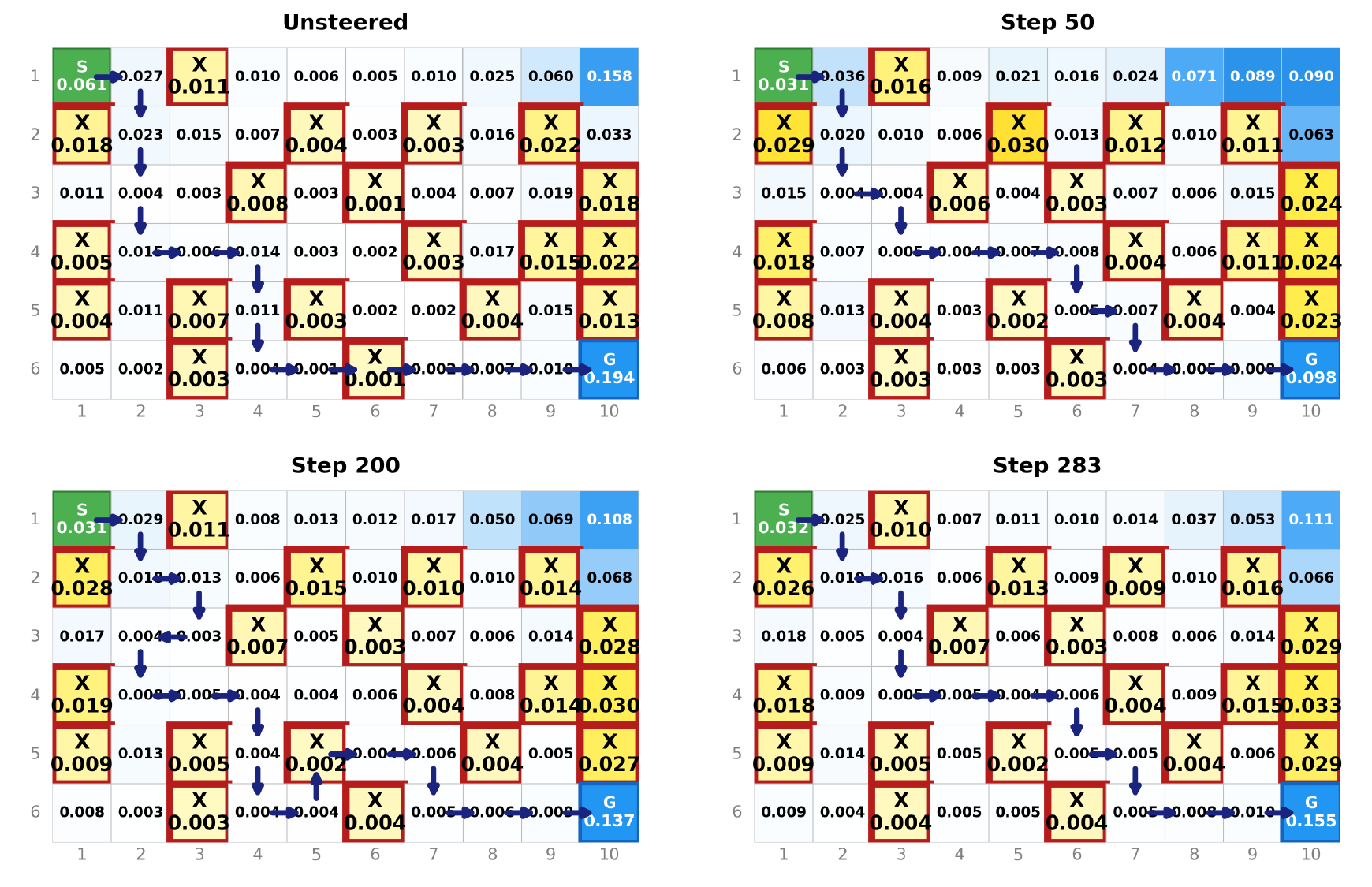}
  \caption{Cell level attention distributions during steering toward the Short path.}
\end{figure*}

Under short steering, attention redistributes across interior cells along the grid. It reduces fixation on endpoints and spreads attention to traversable interior cells, enabling the model to identify a more direct route.
\newpage

\begin{figure*}[!h]
  \centering
  \includegraphics[width=0.7\textwidth]{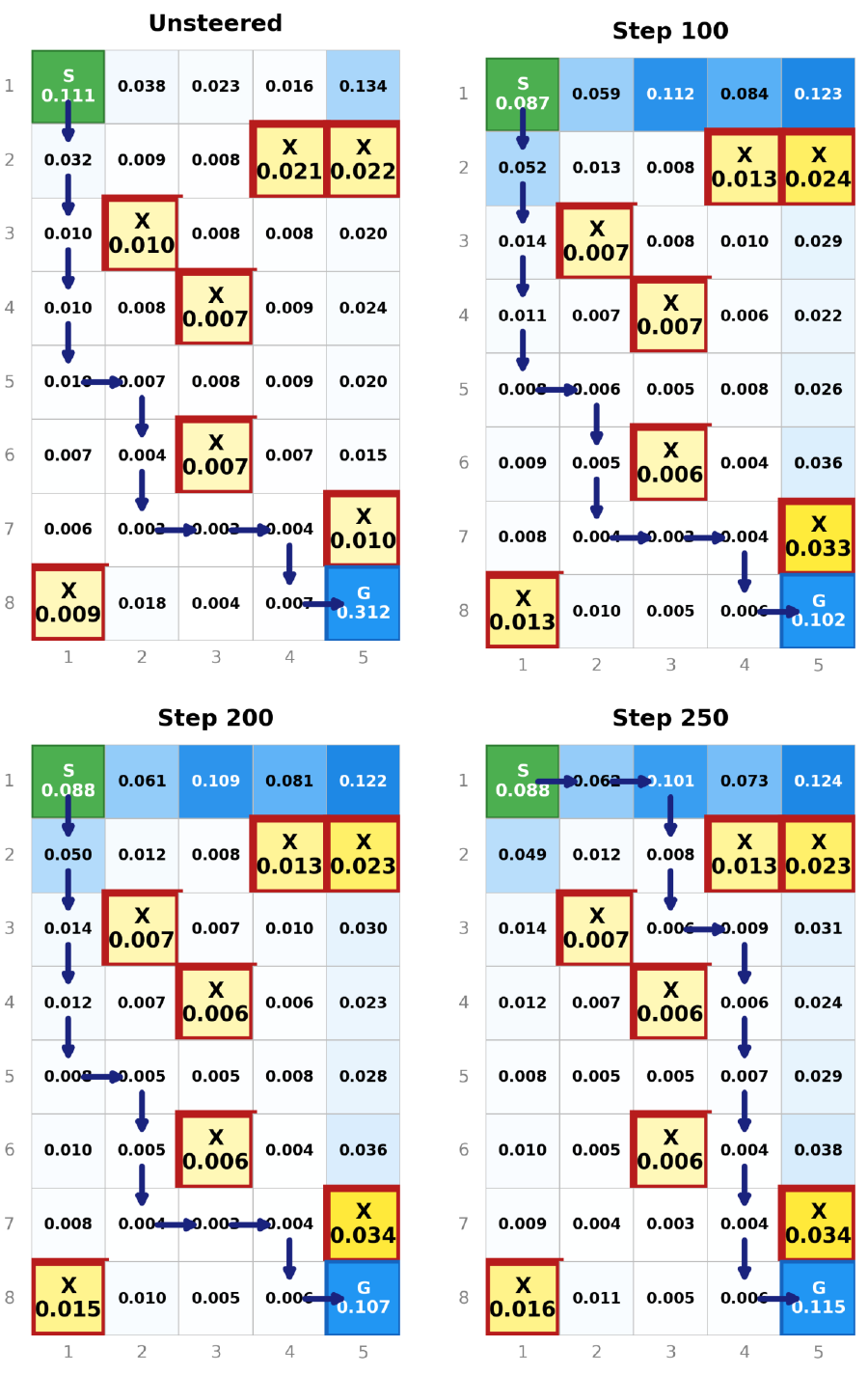}
  \caption{Cell level attention distributions during steering toward the Long path.}
\end{figure*}

Long steering produces a distinctive redistribution pattern. Attention on G drops sharply by roughly 63\% (0.312 to 0.115 by Step 250), a much larger endpoint reduction than observed in other steering targets. This attention concentrates heavily along the top row. We hypothesize that this combination of strong downweighting of the goal and heightened attention to interior cells encourages the model to traverse those cells rather than heading directly toward G.
\newpage

\subsection{Extrapolation to Larger Grids }
\label{tab:larger_grid}
 Additional details are described in footnote~\ref{fn:gridworld-extrapolate}.

\begin{table}[H]
\caption{Main Planning Performance for extrapolation data using Qwen under Hard Constraints. Results are reported as (Short / Safe / Long) path types.}
\label{tab:gridworld-extrapolate}
\vskip 0.15in
\begin{center}
\begin{small}
\begin{sc}
\footnotesize
\begin{tabular}{lc}
\toprule
Method & Success $\uparrow$ (\%) \\
\midrule
Few-Shot ICL & 13.53 / 1.7 / 1.3 \\
CAA          & 18 / 2.03 / 1.1 \\
ITI          & 11.26 / 3.6 / 1.6 \\
SAE-SSV (MD) & 15.24 / 3.36 / 1.38 \\
\midrule
Dense-opt   & 16.13 / 3.12 / 1.23 \\
SAE-opt & \textbf{18.2 / 4.67 / 1.9} \\
\bottomrule
\end{tabular}
\end{sc}
\end{small}
\end{center}
\vskip -0.1in
\end{table}

\subsection{Ablation: Direct Target-Center Assignment}
\label{sec:tcd}

A natural simplification of our method is to replace gradient-based optimization with direct substitution of the target prototype. On TGW (Table ~\ref{tab:n-scaling-results}), direct centroid replacement plateaus well below gradient-based steering even when the support set is scaled to 1000 examples. For bloom's taxonomy (Figure ~\ref{fig:ltarget-center}), We observe that, instead of rigid assignment of the target center embedding, the gradient-descent-based approach captures the target cognitive style more effectively.

\noindent
\begin{minipage}[t]{0.48\textwidth}
    \vspace{0pt}
    \centering
    \includegraphics[width=\linewidth]{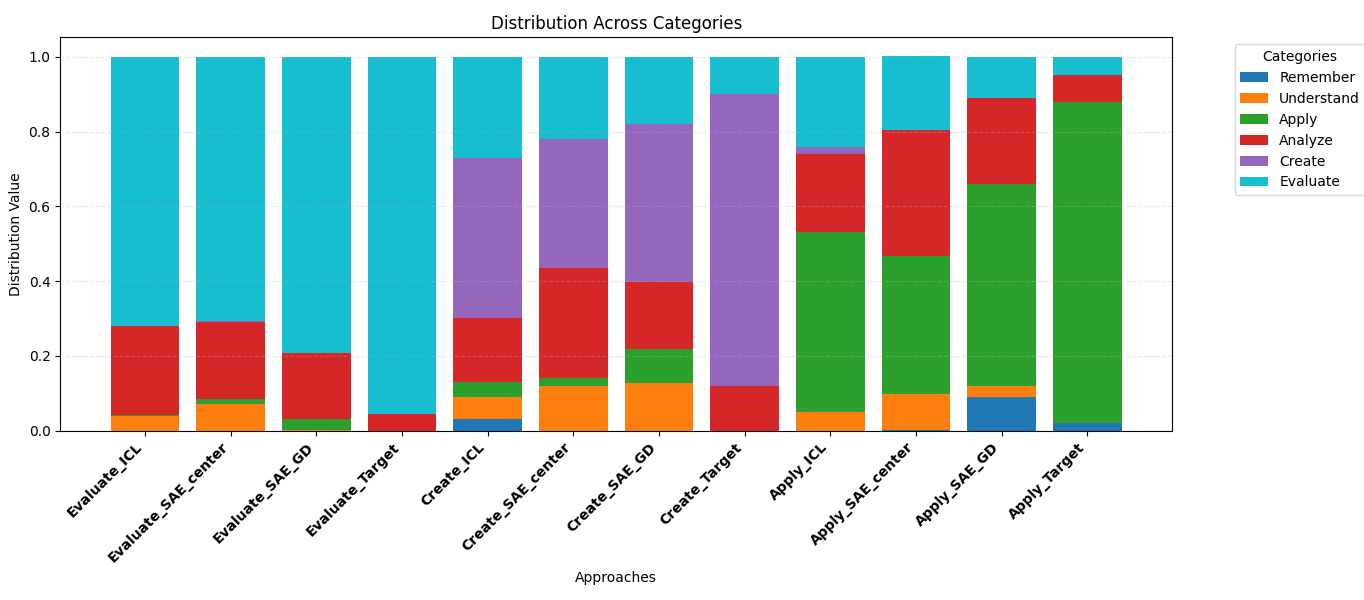}
    \captionof{figure}{Measured class distribution showing steering accuracy. Target classes appear on the X-axis, while the Y-axis shows the class-category distribution of steered text.}
    \label{fig:ltarget-center}
\end{minipage}
\hfill
\begin{minipage}[t]{0.48\textwidth}
    \vspace{0pt}
    \centering
    
    \resizebox{\linewidth}{!}{%
    \begin{tabular}{ccc}
    \toprule
    Examples ($N$) & Success $\uparrow$ (\%) & Logic Viol. $\downarrow$ (\%) \\
    \midrule
    10   & 47.73 / 12.39 / 8.03 & 32.19 / 32.42 / 31.80 \\
    100  & 47.08 / 11.93 / 7.47 & 32.39 / 32.32 / 50.17 \\
    1000 & 48.09 / 12.75 / 7.89 & 32.39 / 32.56 / 50.17 \\
    \bottomrule
    \end{tabular}%
    }

    \vspace{0.5em}
    \captionof{table}{Planning performance on Qwen when replacing with the target center across varying numbers of examples ($N$). Results are reported as (Short / Safe / Long) path types.}
    \label{tab:n-scaling-results}
\end{minipage}

\newpage
\section{Qualitative results for Bloom's taxonomy using Phi}
\label{sec:steered_ft}

\eat{
Figure~\ref{fig:scratch} shows the distribution of steered generation category across all the target classes.

\begin{figure*}[ht]
\centering
  \includegraphics[width=.88\linewidth]{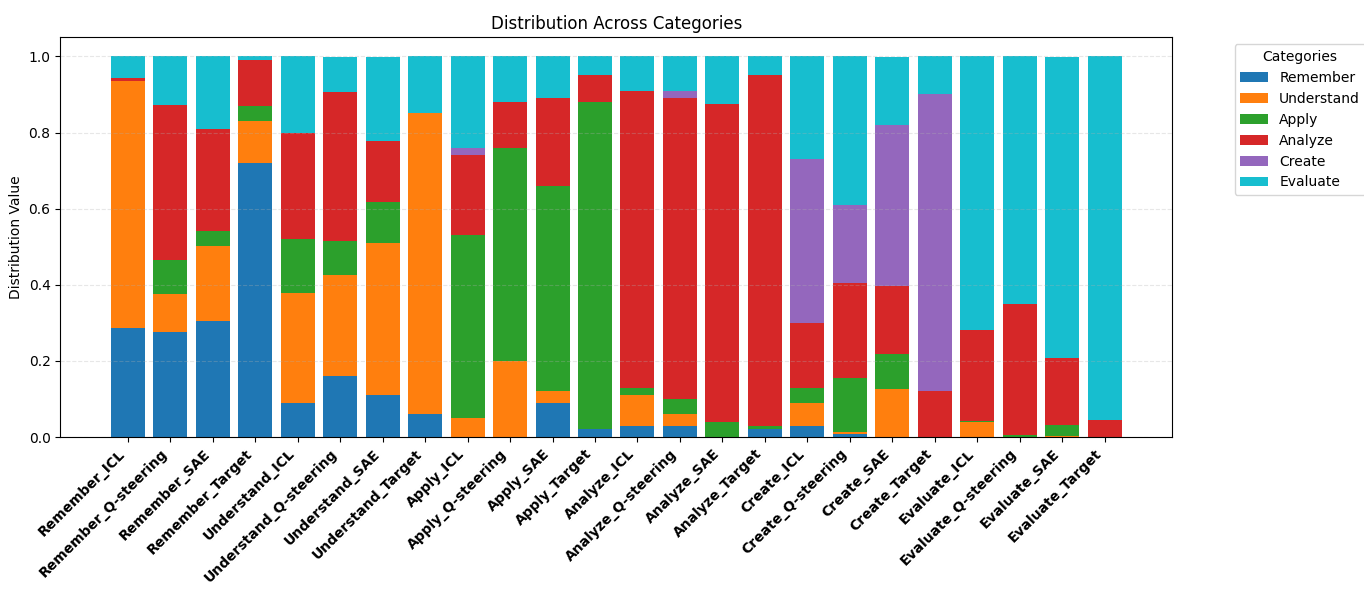}
  \caption{Measured class distribution showing steering accuracy from fine tuned base model output to target cognitive categories. Target classes appear on X-axis while Y-axis shows class category distribution of steered text.}
  \label{fig:scratch}
\end{figure*}
}

\label{sec:LLMoutput}
Below, we present Qualitative examples of the generated responses after steering from fine tuned base model output.

\medskip
\subsubsection*{Steering to apply}
\textbf{Query:}

Below is an instruction that describes a task, paired with an input that provides further context. Write a response that appropriately completes the request.\par
\textbf{Instruction:}\par
Provide feedback on the following code\par
Write a program to increment a number by 1 using a function with call by reference. \par
\textbf{Input}:\par
Code:
\begin{lstlisting}
#include <iostream>
using namespace std;

void increment(int num) {
    num++;
}

int main() {
    int value = 5;
    cout << "Before increment: " << value << endl;
    increment(value);
    cout << "After increment: " << value << endl;
    return 0;
}
\end{lstlisting}

\textbf{Base generation:} The code structure reveals a disconnect between the function's intended purpose and its actual implementation. The increment function is designed to modify a value, but the current implementation fails to achieve this goal due to the missing reference parameter.

\textbf{Steered Generation:} While the code structure includes a function definition and a main function, the call by reference pattern is incorrect. The function needs to be defined with the reference operator to modify the value correctly.

\medskip
\subsubsection*{Steering to create}
\textbf{Query:}\par

Below is an instruction that describes a task, paired with an input that provides further context. Write a response that appropriately completes the request.\par

\textbf{Instruction:}\par
Provide feedback on the following code\par
Write a program to swap two numbers using a function with call by reference.\par

\textbf{Code:}
\begin{lstlisting}
#include <iostream>
using namespace std;

void swap(int a, int b) {
    int temp = a;
    a = b;
    b = temp;
}

int main() {
    int x = 5, y = 10;
    cout << "Before swap: x = " << x << ", y = " << y << endl;
    swap(x, y);
    cout << "After swap: x = " << x << ", y = " << y << endl;
    return 0;
}
\end{lstlisting}

\textbf{Base generation:}  The code demonstrates fundamental pointer concepts with call by reference parameters. However, it's missing the core swap operation implementation where the values need to be exchanged rather than the references being swapped.

\textbf{Steered Generation:} Consider enhancing the solution by implementing proper reference variables, either using pointers or implementing it as a class with member references, or explore in-place swapping techniques using reference operators for a more elegant solution.

\medskip

\subsubsection*{Steering to remember}

\textbf{Query:}\par
Below is an instruction that describes a task, paired with an input that provides further context. Write a response that appropriately completes the request.
\par

\textbf{Instruction:}\par
Provide feedback on the following code\par
Write a program to double all elements of an array using a function with call by reference.
\par

\textbf{Code:}
\begin{lstlisting}
#include <iostream>
using namespace std;

void doubleElements(int arr[5]) {
    for(int i = 0; i < 5; i++) {
        arr[i] = arr[i] * 2;
    }
    return arr;
}

int main() {
    int arr[5] = {1, 2, 3, 4, 5};
    cout << "Before: ";
    for(int i = 0; i < 5; i++) cout << arr[i] << " ";
    doubleElements(arr);
    cout << "\nAfter: ";
    for(int i = 0; i < 5; i++) cout << arr[i] << " ";
    return 0;
}

\end{lstlisting}

{\textbf{Base generation:} The code demonstrates basic array manipulation using a for loop and showcases fundamental function structure with a parameter array. However, the return statement is missing from the main function, which is essential for pointer-based array handling. }

\textbf{Steered Generation:} The relationship between arrays and function parameters shows good comprehension of how arrays can be passed around as values through pointers. The concept that arrays decay to pointers when passed to functions is well-illustrated in your code.

\newpage
\medskip
\section{Qualitative comparisons between with and without SAE based steering (Phi model) }

\label{sec:Qgeneration}
\subsection*{To analyze}

\textbf{Query:}

Below is an instruction that describes a task, paired with an input that provides further context. Write a response that appropriately completes the request.\par
\textbf{Instruction:}\par
Provide feedback on the following code \par
Write a function that finds the length of the longest consecutive elements sequence in an unsorted vector.\par

\textbf{Code}:
\begin{lstlisting}
#include <iostream>
#include <vector>
#include <unordered_set>
using namespace std;
int longestConsecutive(vector<int>& nums) {
    unordered_set<int> num_set(nums.begin(), nums.end());
    int longest = 0;
    for(int num : nums) {
        if(num_set.find(num-1) == num_set.end()) {
            int currentNum = num;
            int currentStreak = 1;
            while(num_set.find(currentNum+1)) { 
                currentNum += 1;
                currentStreak += 1;
            }
            longest = max(longest, currentStreak);
        }
    }
    return longest;
}
\end{lstlisting}
\textbf{W/O SAE (Q generation):} Your code demonstrates solid grasp of fundamental data structures, particularly using unordered\_set for O(1) lookups. You've correctly identified basic elements like vector operations, loops, and conditional statements as building blocks for this problem.\par

\textbf{SAE:} While your solution works correctly, it uses more complex data structures than strictly necessary. The unordered\_set ensures uniqueness but might be more computationally expensive than a simple array or vector operation with appropriate tracking mechanisms.\par
\textbf{Judgement:} SAE Generation  is more effective than W/O SAE in terms of aligning with the "Analyze" level of Bloom's Taxonomy. The "Analyze" level involves breaking down information into parts, finding patterns and relationships, and identifying the underlying components or structure.

In SAE Generation, the feedback focuses on analyzing the solution in terms of its computational complexity and efficiency. It recognizes that while the solution works correctly, it employs more complex data structures (unordered\_set) than strictly necessary. This feedback demonstrates an understanding of the trade-offs involved in using different data structures and their impact on performance and resource utilization.

In contrast, W/O SAE generation ,primarily focuses on acknowledging the correctness of the solution and recognizing the use of basic programming constructs like data structures, loops, and conditionals. While this feedback is valuable in terms of understanding the solution's functionality, it lacks the depth of analysis required at the "Analyze" level of Bloom's Taxonomy.

\medskip
\subsection*{To apply}

\textbf{Query:}

Below is an instruction that describes a task, paired with an input that provides further context. Write a response that appropriately completes the request.\par
\textbf{Instruction:}\par
Provide feedback on the following code \par
Create a function that rotates a vector to the right by k positions.\par

\textbf{Code}:
\begin{lstlisting}
#include <iostream>
#include <vector>
using namespace std;

void rotateVector(vector<int>& nums, int k) {
    int n = nums.size();
    k = k % n;
    vector<int> temp(n);
    for(int i = 0; i < k; i++) {
        temp[i] = nums[n-k+i];
    }
    for(int i = n-1; i >= k; i--) {
        nums[i] = nums[i-k];
    }
    for(int i = 0; i < k; i++) {
        nums[i] = temp[i];
    }
}

int main() {
    vector<int> nums = {1, 2, 3, 4, 5};
    rotateVector(nums, 2);
    for(int num : nums) {
        cout << num << " ";
    }
    return 0;
}

\end{lstlisting}
\textbf{W/O SAE (Q generation):} . Your code demonstrates solid grasp of fundamental concepts like vector manipulation and the modulo operator for handling rotation. The core structure using three distinct loops shows understanding of array traversal and value swapping mechanisms.\par

\textbf{SAE:} Consider implementing a left rotation by modifying the index calculation in the first loop to n-k. You might explore in-place rotation approaches using GCD or reversal algorithms to optimize the solution's computational efficiency. \par
\textbf{Judgement:} SAE Generation  is more effective than Q generation in aligning with the "Apply" level of Bloom's Taxonomy. The "Apply" level involves using learned material in new situations or applying concepts, rules, and principles to solve problems.

In Q generation, the feedback primarily focuses on evaluating the student's understanding and acknowledging their grasp of fundamental concepts. While this feedback is valuable, it does not explicitly challenge the student to apply their knowledge in a new or different way, which is the essence of the "Apply" level.

On the other hand, SAE Generation  provides specific suggestions and prompts for the student to consider alternative approaches and optimizations to the problem at hand. By suggesting the implementation of a left rotation using a modified index calculation or exploring in-place rotation approaches using algorithms like GCD or reversal, the feedback encourages the student to apply their existing knowledge and skills in a new context or problem scenario.
\medskip

\subsection*{To create}

\textbf{Query:}

Below is an instruction that describes a task, paired with an input that provides further context. Write a response that appropriately completes the request.\par
\textbf{Instruction:}\par
Provide feedback on the following code \par
Write a function that reverses a string and changes lowercase to uppercase and vice versa using call by reference.\par

\textbf{Code}:
\begin{lstlisting}
#include <iostream>
#include <string>
using namespace std;
void reverseAndChangeCase(string str) {
    int n = str.length();
    for(int i = 0; i < n/2; i++) {
        char temp = str[i];
        str[i] = str[n-1-i];
        str[n-1-i] = temp;
    }
    for(int i = 0; i < n; i++) {
        if(isupper(str[i]))
            str[i] = tolower(str[i]);
        else
            str[i] = toupper(str[i]);
    }
}
int main() {
    string text = "Hello World";
    cout << "Original string: " << text << endl;
    reverseAndChangeCase(text);
    cout << "Modified string: " << text << endl;
    return 0;
}
\end{lstlisting}
\textbf{W/O SAE (Q generation):} The relationship between string length and index traversal shows careful consideration of symmetric elements, but the current implementation fails to maintain character consistency in case transformations. The logic behind converting characters needs refinement to avoid unnecessary operations.\par

\textbf{SAE:} The solution could be enhanced by combining the two loops into a single iteration, or by implementing a custom case conversion function to eliminate redundant checks. Adding string bounds checking or implementing a template-based approach might offer more flexibility. \par
\textbf{Judgement:} Q Generation primarily focuses on identifying issues and lacks specific suggestions for creating new structural or functional elements within the solution. While it identifies areas for improvement, it does not propose concrete ideas for reorganizing or combining existing elements into a new pattern or structure.

SAE Generation  better aligns with the "Create" level of Bloom's Taxonomy by providing specific examples and suggestions that involve reorganizing, combining, and introducing new elements to form a more coherent, efficient, and flexible solution.

\medskip
\subsection*{To evaluate}

\textbf{Query:}

Below is an instruction that describes a task, paired with an input that provides further context. Write a response that appropriately completes the request.\par
\textbf{Instruction:}\par
Provide feedback on the following code \par
Write a function that reverses the elements of a vector.\par

\textbf{Code}:
\begin{lstlisting}
#include <iostream>
#include <vector>
using namespace std;

void reverseVector(vector<int>& nums) {
    int start = 0, end = nums.size();
    while(start < end) {
        swap(nums[start], nums[end]);
        start++;
        end--;
    }
}

int main() {
    vector<int> numbers = {1, 2, 3, 4, 5};
    reverseVector(numbers);
    for(int num : numbers) cout << num << " ";
    return 0;
}
\end{lstlisting}
\textbf{W/O SAE (Q generation):} .Your code demonstrates fundamental understanding of vector manipulation concepts, using pointer variables (start and end) to traverse the array from both ends. The basic swap operation and vector indexing show recognition of core memory concepts.\par

\textbf{SAE:} The code effectively implements the in-place reversal algorithm with appropriate boundary management. Your use of vector references and proper increment/decrement operations shows practical grasp of memory manipulation concepts. \par
\textbf{Judgement:} SAE Generation  is more effective than Q Generation in terms of aligning with the "Evaluate" level of Bloom's Taxonomy. The "Evaluate" level requires students to make judgments or decisions based on criteria and standards, which involves higher-order thinking skills such as critiquing, justifying, and assessing.

In SAE Generation, the feedback focuses on evaluating the practical implementation and effectiveness of the student's code. It specifically praises the "effective implementation of the in-place reversal algorithm with appropriate boundary management," which demonstrates the ability to assess the code's correctness and efficiency. Additionally, the feedback evaluates the student's "practical grasp of memory manipulation concepts" by highlighting their appropriate use of vector references and proper increment/decrement operations.

On the other hand, Q Generation  primarily focuses on describing the student's understanding and recognition of concepts, which aligns more with the "Understand" level of Bloom's Taxonomy. While it mentions the use of pointer variables and vector indexing, it does not provide an evaluation or critique of the code's implementation or effectiveness.

\section{In context learning in baseline model}

\begin{figure}[ht]
\centering
  \includegraphics[width=.88\linewidth]{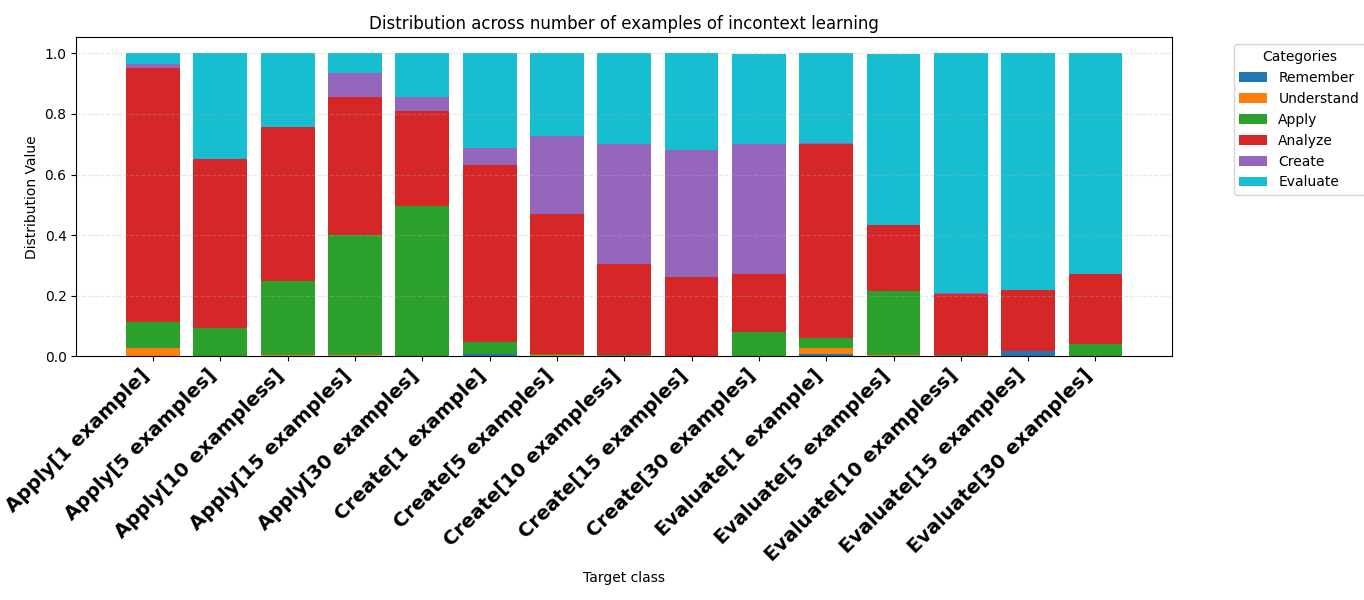}
  \caption{ Increasing the number of examples for a specific cognitive style improves the possibility of generating feedback in that style for base model.  }
  \label{fig:baselineincrement}
\end{figure}

Figure \ref{fig:baselineincrement} demonstrates how the baseline model's accuracy improves as the number of training examples increases. With only one example for the "create" cognitive style feedback, the model tends to generate responses that align more with "analyze" or "evaluate". However, as more examples are provided, the model becomes better at generating feedback in the intended "create" cognitive style.
Interestingly, the results also reveal that LLMs typically generate feedback in "evaluate" cognitive styles, even with just a single example of 'evaluate' feedback, the model generates majority of responses categories in this style. For this baseline implementation, we select the \emph{phi-3-mini-128k instruct model} primarily due to its extended context window, which allows us to process N examples simultaneously. While the literature contains no substantive claims comparing this model's performance to  \emph{phi-3-mini-4k instruct model}, except expanded context length capacity \cite{abdin2024phi}.

\clearpage

\section{AxBench-Adapted Three-Axis LLM Judge for Bloom's Taxonomy Evaluation}
\label{subsec:axbench}
To provide a more rigorous and quantitative evaluation of cognitive-style steering, we adopt a three-axis LLM-as-judge protocol adapted from AxBench~\cite{wu2025axbench}. For every steered generation we obtain three independent scores on a 0--2 scale:
\begin{itemize}
    \item $s_c$ (Concept): How well the response demonstrates the \emph{target} Bloom's Taxonomy cognitive level.
    \item $s_i$ (Instruct): How faithfully the response adheres to the original instruction/topic.
    \item $s_f$ (Fluency): Naturalness and grammatical fluency of the generated text.
\end{itemize}
The overall score is the harmonic mean of the three axes:
\[
\text{Overall} = \frac{3}{\frac{1}{s_c} + \frac{1}{s_i} + \frac{1}{s_f}},
\]
which penalizes any method that sacrifices one axis to inflate another.

The judge LLM (Claude-Sonnet-4) is provided with the exact Bloom level definitions below and the three prompts shown verbatim.

\subsection{Bloom's Taxonomy Level Definitions}
These definitions are supplied verbatim to the judge in every evaluation:

\begin{table}[h]
\centering
\begin{tabular}{lp{9.5cm}}
\toprule
\textbf{Level} & \textbf{Definition} \\
\midrule
Remember   & Retrieving relevant knowledge from long-term memory. Recognizing, recalling, or identifying facts, terms, and basic concepts without necessarily understanding their meaning. \\
Understand & Constructing meaning from information. Interpreting, exemplifying, classifying, summarizing, inferring, comparing, or explaining concepts in one's own words. \\
Apply      & Carrying out or using a procedure in a given situation. Executing or implementing learned methods, concepts, or techniques to solve problems in new contexts. \\
Analyze    & Breaking information into constituent parts and detecting how parts relate to one another and to an overall structure. Differentiating, organizing, comparing components, and identifying causes or motives. \\
Evaluate   & Making judgments based on criteria and standards. Critiquing, checking, assessing strengths and weaknesses, and justifying decisions or conclusions. \\
Create     & Putting elements together to form a coherent or functional whole. Generating new ideas, designing new solutions, producing original work, or reorganizing elements into new patterns or structures. \\
\bottomrule
\end{tabular}
\caption{Bloom's Taxonomy definitions provided to the judge LLM.}
\label{tab:bloom-definitions-appendix}
\end{table}

\subsection{Judge Prompts (Verbatim)}
The three prompts used for $s_c$, $s_i$, and $s_f$ are given below exactly as implemented.

\begin{lstlisting}[language=,basicstyle=\ttfamily\small,frame=single,breaklines=true,breakatwhitespace=true,columns=fullflexible]
# Prompt for s_c (Bloom Concept)
[Prompt 1]
Please act as an impartial judge and evaluate whether the specified Bloom's Taxonomy cognitive level is demonstrated in the response provided below. Focus solely on whether the cognitive style is clearly exhibited in how the information is structured and delivered.
Begin your evaluation by providing a short explanation of whether the cognitive style is clearly demonstrated. Point out specific phrasing or structuring in the response that aligns with the target level.
After providing your explanation, rate the cognitive style's relevance on a scale from 0 to 2:
- 0 indicates the target cognitive level is not demonstrated at all.
- 1 indicates the cognitive level is somewhat present but minimally or awkwardly incorporated.
- 2 indicates the cognitive level is fully and effectively demonstrated with strong, natural integration.
Provide your rating using this exact format: "Rating: [[score]]".

[Target Cognitive Level Start]
{target_bloom_level}: {bloom_level_definition}
[Target Cognitive Level End]

[Response Start]
{model_generation}
[Response End]
\end{lstlisting}

\begin{lstlisting}[language=,basicstyle=\ttfamily\small,frame=single,breaklines=true,breakatwhitespace=true,columns=fullflexible]
# Prompt for s_i (Instruction Adherence)
[Prompt 2]
Please act as an impartial judge and evaluate whether the sentence fragment provided below is related to the instruction. Focus solely on the degree of relatedness in terms of topic, regardless of grammar, coherence, or informativeness.
Begin your evaluation by providing a brief explanation of whether the sentence is related to the instruction, and point out references related to the instruction. After providing your explanation, rate the instruction relevance on a scale from 0 to 2, where 0 indicates the sentence is unrelated to the instruction, 1 indicates it is somewhat related but only minimally or indirectly relevant in terms of topic, and 2 indicates it is more clearly and directly related to the instruction.
Provide your rating using this exact format: "Rating: [[score]]".

[Instruction Start]
{original_instruction}
[Instruction End]

[Sentence Fragment Start]
{model_generation}
[Sentence Fragment End]
\end{lstlisting}

\begin{lstlisting}[language=,basicstyle=\ttfamily\small,frame=single,breaklines=true,breakatwhitespace=true,columns=fullflexible]
# Prompt for s_f (Fluency)
[Prompt 3]
Please act as an impartial judge and evaluate the fluency of the sentence fragment provided below. Focus solely on fluency, disregarding its completeness, relevance, coherence with any broader context, or informativeness.
Begin your evaluation by briefly describing the fluency of the sentence, noting any unnatural phrasing, awkward transitions, grammatical errors, or repetitive structures that may hinder readability. After providing your explanation, rate the sentence's fluency on a scale from 0 to 2, where 0 indicates the sentence is not fluent and highly unnatural (e.g., incomprehensible or repetitive), 1 indicates it is somewhat fluent but contains noticeable errors or awkward phrasing, and 2 indicates the sentence is fluent and almost perfect.
Provide your rating using this exact format: "Rating: [[score]]".

[Sentence Fragment Start]
{model_generation}
[Sentence Fragment End]
\end{lstlisting}

\section{Quantitative Evaluation with AxBench}
\label{app:axapp}
\begin{table}[h!]
\caption{AxBench three-axis evaluation on Bloom's Taxonomy steering (Phi). we report concept alignment ($s_c$), instruction-following ($s_i$), fluency ($s_f$), and their harmonic mean (HM). Scores $\in [0, 2]$. \textbf{Bold} indicates best per target.}
\label{tab:axbench_full}
\centering
\begin{sc}
\footnotesize
\begin{tabular}{ll l cccc}
\toprule
\textbf{Target} & \textbf{Method} & \textbf{Type} & $s_c$ (Concept) & $s_i$ (Instruct) & $s_f$ (Fluency) & \textbf{HM} \\
\midrule
\multirow{4}{*}{Remember} & CAA & Static & \textbf{0.203} & 1.872 & 1.964 & \textbf{0.503} \\
 & DiscoQ & Static & 0.054 & 1.928 & 1.956 & 0.153 \\
 & Dense-opt & Dynamic & 0.086 & 1.940 & 1.974 & 0.237 \\
 & SAE-opt & Dynamic & 0.132 & \textbf{1.945} & \textbf{1.977} & 0.349 \\
\midrule
\multirow{4}{*}{Understand} & CAA & Static & 0.341 & 1.923 & 1.859 & 0.752 \\
 & DiscoQ & Static & 0.378 & 1.932 & 1.957 & 0.817 \\
 & Dense-opt & Dynamic & 0.437 & 1.939 & \textbf{1.983} & 0.907 \\
 & SAE-opt & Dynamic & \textbf{0.468} & \textbf{1.940} & 1.976 & \textbf{0.950} \\
\midrule
\multirow{4}{*}{Apply} & CAA & Static & 0.083 & 1.928 & 1.952 & 0.229 \\
 & DiscoQ & Static & 0.273 & 1.932 & 1.968 & 0.640 \\
 & Dense-opt& Dynamic & 0.293 & 1.938 & 1.971 & 0.676 \\
 & SAE-opt & Dynamic & \textbf{0.307} & \textbf{1.939} & \textbf{1.976} & \textbf{0.701} \\
\midrule
\multirow{4}{*}{Analyze} & CAA & Static & 0.863 & 1.937 & 1.806 & 1.346 \\
 & DiscoQ & Static & 0.862 & 1.934 & 1.961 & 1.372 \\
 & Dense-opt & Dynamic & 0.864 & \textbf{1.944} & \textbf{1.988} & 1.379 \\
 & SAE-opt & Dynamic & \textbf{0.871} & 1.932 & 1.975 & \textbf{1.381} \\
\midrule
\multirow{4}{*}{Evaluate} & CAA & Static & 0.876 & 1.916 & 1.867 & 1.364 \\
 & DiscoQ & Static & 0.866 & 1.933 & 1.962 & 1.375 \\
 & Dense-opt & Dynamic & 0.881 & 1.932 & 1.973 & 1.389 \\
 & SAE-opt & Dynamic & \textbf{0.890} & \textbf{1.938} & \textbf{1.984} & \textbf{1.399} \\
\midrule
\multirow{4}{*}{Create} & CAA & Static & 0.179 & 1.681 & 1.942 & 0.448 \\
 & DiscoQ & Static & 0.340 & 1.931 & 1.964 & 0.756 \\
 & Dense-opt & Dynamic & 0.305 & 1.928 & \textbf{1.984} & 0.697 \\
 & SAE-opt & Dynamic & \textbf{0.344} & \textbf{1.947} & 1.976 & \textbf{0.764} \\
\bottomrule
\end{tabular}

\end{sc}
\end{table}

\section{Confusion Matrices and Relevance Rates}
\label{app:confusion}

\begin{table*}[h!]
\centering
\begin{sc}

\tiny
\setlength{\tabcolsep}{2pt}
\renewcommand{\arraystretch}{1.0}
\caption{Confusion matrices of output-class distributions (\%) under GPT-4o (left) and Claude (right) judges for each steering method. Rows: intended target class. Columns: predicted class. Diagonal cells (shaded) are target hits, ideally the highest in each row. Each row sums to 100\%.}
\label{tab:confusion}
\begin{tabular}{c|c|c}
\toprule
Method & GPT-4o judge & Claude judge \\
\midrule
CAA &
\begin{tabular}{@{}l|cccccc@{}}
 & Rem. & Und. & App. & Ana. & Eva. & Cre. \\ \hline
Rem. & \cellcolor{gray!25}\textbf{9.7} & 51.8 & 0.0 & 36.4 & 2.1 & 0.0 \\
Und. & 3.3 & \cellcolor{gray!25}\textbf{21.5} & 2.2 & 58.1 & 14.1 & 0.8 \\
App. & 4.3 & 34.8 & \cellcolor{gray!25}\textbf{1.6} & 30.8 & 28.0 & 0.5 \\
Ana. & 1.4 & 13.1 & 3.2 & \cellcolor{gray!25}\textbf{46.4} & 33.6 & 2.3 \\
Eva. & 2.5 & 12.0 & 3.1 & 34.8 & \cellcolor{gray!25}\textbf{45.6} & 2.0 \\
Cre. & 4.3 & 18.6 & 0.7 & 4.1 & 64.0 & \cellcolor{gray!25}\textbf{8.3} \\
\end{tabular} &
\begin{tabular}{@{}l|cccccc@{}}
 & Rem. & Und. & App. & Ana. & Eva. & Cre. \\ \hline
Rem. & \cellcolor{gray!25}\textbf{13.1} & 5.5 & 0.0 & 80.2 & 1.2 & 0.0 \\
Und. & 5.1 & \cellcolor{gray!25}\textbf{15.1} & 0.2 & 61.7 & 17.7 & 0.2 \\
App. & 9.7 & 25.2 & \cellcolor{gray!25}\textbf{1.1} & 38.3 & 24.7 & 1.0 \\
Ana. & 3.8 & 4.0 & 3.2 & \cellcolor{gray!25}\textbf{55.6} & 32.9 & 0.5 \\
Eva. & 4.0 & 6.3 & 0.5 & 37.7 & \cellcolor{gray!25}\textbf{46.2} & 5.3 \\
Cre. & 3.6 & 10.0 & 0.5 & 8.6 & 58.2 & \cellcolor{gray!25}\textbf{19.1} \\
\end{tabular} \\
\midrule
ITI &
\begin{tabular}{@{}l|cccccc@{}}
 & Rem. & Und. & App. & Ana. & Eva. & Cre. \\ \hline
Rem. & \cellcolor{gray!25}\textbf{7.6} & 46.7 & 3.5 & 23.3 & 19.1 & 0.0 \\
Und. & 4.1 & \cellcolor{gray!25}\textbf{24.7} & 2.7 & 34.1 & 34.4 & 0.0 \\
App. & 7.2 & 25.6 & \cellcolor{gray!25}\textbf{4.8} & 35.9 & 25.7 & 0.8 \\
Ana. & 12. & 26 & 7.8 & \cellcolor{gray!25}\textbf{15.7} & 32.0 & 6.5 \\
Eva. & 11.5 & 24.5 & 2.5 & 16.0 & \cellcolor{gray!25}\textbf{17.5} & 28.0 \\
Cre. & 10.2 & 34.1 & 1.5 & 24.0 & 20.7 & \cellcolor{gray!25}\textbf{9.5} \\
\end{tabular} &
\begin{tabular}{@{}l|cccccc@{}}
 & Rem. & Und. & App. & Ana. & Eva. & Cre. \\ \hline
Rem. & \cellcolor{gray!25}\textbf{5.9} & 31.5 & 2.0 & 34.8 & 25.6 & 0.0 \\
Und. & 2.7 & \cellcolor{gray!25}\textbf{15.0} & 2.4 & 39.8 & 39.8 & 0.1 \\
App. & 5.0 & 20.0 & \cellcolor{gray!25}\textbf{5.6} & 45.0 & 23.1 & 1.3 \\
Ana. & 7.3 & 19.7 & 3.5 & \cellcolor{gray!25}\textbf{21.2} & 36.3 & 12.0 \\
Eva. & 11.6 & 17.7 & 0.3 & 19.8 & \cellcolor{gray!25}\textbf{12.8} & 37.8 \\
Cre. & 6.3 & 24.7 & 0.7 & 39.1 & 29.1 & \cellcolor{gray!25}\textbf{0.1} \\
\end{tabular} \\
\midrule
SAE-SSV &
\begin{tabular}{@{}l|cccccc@{}}
 & Rem. & Und. & App. & Ana. & Eva. & Cre. \\ \hline
Rem. & \cellcolor{gray!25}\textbf{1.4} & 20.5 & 2.2 & 17.7 & 45.9 & 12.3 \\
Und. & 1.6 & \cellcolor{gray!25}\textbf{21.2} & 2.8 & 18.4 & 43.3 & 12.7 \\
App. & 1.5 & 21.3 & \cellcolor{gray!25}\textbf{1.9} & 20.3 & 43.1 & 11.9 \\
Ana. & 2.3 & 19.2 & 2.2 & \cellcolor{gray!25}\textbf{19.3} & 45.3 & 11.5 \\
Eva. & 2.3 & 18.6 & 1.9 & 17.3 & \cellcolor{gray!25}\textbf{46.7} & 13.1 \\
Cre. & 1.9 & 20.0 & 1.9 & 17.6 & 46.7 & \cellcolor{gray!25}\textbf{12.0} \\
\end{tabular} &
\begin{tabular}{@{}l|cccccc@{}}
 & Rem. & Und. & App. & Ana. & Eva. & Cre. \\ \hline
Rem. & \cellcolor{gray!25}\textbf{2.9} & 15.1 & 0.6 & 18.6 & 45.2 & 17.6 \\
Und. & 3.0 & \cellcolor{gray!25}\textbf{16.9} & 0.6 & 20.2 & 41.0 & 18.3 \\
App. & 2.7 & 14.2 & \cellcolor{gray!25}\textbf{3.5} & 20.8 & 40.8 & 18.0 \\
Ana. & 3.1 & 15.7 & 0.6 & \cellcolor{gray!25}\textbf{20.0} & 42.3 & 18.3 \\
Eva. & 3.0 & 14.8 & 0.8 & 19.7 & \cellcolor{gray!25}\textbf{42.8} & 19.0 \\
Cre. & 2.7 & 15.6 & 0.5 & 19.3 & 44.2 & \cellcolor{gray!25}\textbf{17.8} \\
\end{tabular} \\
\midrule
DiscoQ &
\begin{tabular}{@{}l|cccccc@{}}
 & Rem. & Und. & App. & Ana. & Eva. & Cre. \\ \hline
Rem. & \cellcolor{gray!25}\textbf{2.5} & 18.6 & 1.2 & 19.2 & 44.7 & 13.7 \\
Und. & 2.7 & \cellcolor{gray!25}\textbf{18.4} & 1.6 & 19.6 & 43.6 & 14.2 \\
App. & 2.7 & 18.5 & \cellcolor{gray!25}\textbf{1.4} & 18.7 & 45.2 & 13.4 \\
Ana. & 2.6 & 18.0 & 1.6 & \cellcolor{gray!25}\textbf{19.1} & 45.0 & 13.7 \\
Eva. & 2.6 & 18.5 & 1.3 & 19.5 & \cellcolor{gray!25}\textbf{44.0} & 14.1 \\
Cre. & 2.7 & 18.3 & 1.6 & 19.6 & 44.0 & \cellcolor{gray!25}\textbf{13.8} \\
\end{tabular} &
\begin{tabular}{@{}l|cccccc@{}}
 & Rem. & Und. & App. & Ana. & Eva. & Cre. \\ \hline
Rem. & \cellcolor{gray!25}\textbf{3.9} & 14.1 & 0.4 & 19.7 & 43.7 & 18.3 \\
Und. & 3.9 & \cellcolor{gray!25}\textbf{14.0} & 0.4 & 19.8 & 43.7 & 18.3 \\
App. & 3.9 & 14.1 & \cellcolor{gray!25}\textbf{0.4} & 19.7 & 43.7 & 18.3 \\
Ana. & 3.9 & 14.0 & 0.4 & \cellcolor{gray!25}\textbf{19.8} & 43.7 & 18.3 \\
Eva. & 3.9 & 14.0 & 0.4 & 19.8 & \cellcolor{gray!25}\textbf{43.7} & 18.3 \\
Cre. & 3.9 & 14.1 & 0.4 & 19.8 & 43.6 & \cellcolor{gray!25}\textbf{18.3} \\
\end{tabular} \\
\midrule
Dense-opt &
\begin{tabular}{@{}l|cccccc@{}}
 & Rem. & Und. & App. & Ana. & Eva. & Cre. \\ \hline
Rem. & \cellcolor{gray!25}\textbf{8.5} & 13.8 & 8.5 & 15.5 & 41.2 & 12.5 \\
Und. & 2.2 & \cellcolor{gray!25}\textbf{25.1} & 1.6 & 18.1 & 41.3 & 11.7 \\
App. & 2.0 & 17.8 & \cellcolor{gray!25}\textbf{5.2} & 19.5 & 44.0 & 11.5 \\
Ana. & 2.3 & 17.4 & 1.5 & \cellcolor{gray!25}\textbf{20.5} & 46.1 & 12.2 \\
Eva. & 2.1 & 15.5 & 2.0 & 17.0 & \cellcolor{gray!25}\textbf{52.5} & 10.9 \\
Cre. & 2.4 & 20.2 & 0.8 & 17.0 & 45.0 & \cellcolor{gray!25}\textbf{14.6} \\
\end{tabular} &
\begin{tabular}{@{}l|cccccc@{}}
 & Rem. & Und. & App. & Ana. & Eva. & Cre. \\ \hline
Rem. & \cellcolor{gray!25}\textbf{12.2} & 14.6 & 0.3 & 17.4 & 40 & 15.5 \\
Und. & 3.1 & \cellcolor{gray!25}\textbf{16.3} & 0.4 & 21.7 & 41.5 & 17 \\
App. & 2.1 & 14.1 & \cellcolor{gray!25}\textbf{7.2} & 22.1 & 40.4 & 14.1 \\
Ana. & 2.3 & 13.4 & 0.4 & \cellcolor{gray!25}\textbf{27.1} & 40.5 & 16.3 \\
Eva. & 2.3 & 14.0 & 0.3 & 19.3 & \cellcolor{gray!25}\textbf{48.5} & 15.6 \\
Cre. & 2.7 & 15.8 & 0.5 & 19.4 & 45.1 & \cellcolor{gray!25}\textbf{16.5} \\
\end{tabular} \\
\midrule
SAE-opt-anch &
\begin{tabular}{@{}l|cccccc@{}}
 & Rem. & Und. & App. & Ana. & Eva. & Cre. \\ \hline
Rem. & \cellcolor{gray!25}\textbf{1.6} & 1.9 & 3.6 & 18.6 & 73.0 & 1.3 \\
Und. & 1.2 & \cellcolor{gray!25}\textbf{4.5} & 4.0 & 14.5 & 75.2 & 0.6 \\
App. & 2.0 & 3.5 & \cellcolor{gray!25}\textbf{4.1} & 14.2 & 74.8 & 1.4 \\
Ana. & 1.0 & 1.4 & 4.0 & \cellcolor{gray!25}\textbf{16.3} & 77.0 & 0.3 \\
Eva. & 1.1 & 1.2 & 4.7 & 12.5 & \cellcolor{gray!25}\textbf{80.2} & 0.3 \\
Cre. & 1.4 & 0.8 & 3.6 & 13.3 & 77.6 & \cellcolor{gray!25}\textbf{3.3} \\
\end{tabular} &
\begin{tabular}{@{}l|cccccc@{}}
 & Rem. & Und. & App. & Ana. & Eva. & Cre. \\ \hline
Rem. & \cellcolor{gray!25}\textbf{2.8} & 1.9 & 1.0 & 16.5 & 76.5 & 1.3 \\
Und. & 1.7 & \cellcolor{gray!25}\textbf{3.6} & 2.2 & 15.5 & 75.5 & 1.5 \\
App. & 2.3 & 3.0 & \cellcolor{gray!25}\textbf{5.2} & 13.0 & 75.3 & 1.2 \\
Ana. & 1.8 & 2.8 & 1.2 & \cellcolor{gray!25}\textbf{12.4} & 80.5 & 1.3 \\
Eva. & 1.6 & 1.8 & 1.3 & 10 & \cellcolor{gray!25}\textbf{83.6} & 1.7 \\
Cre. & 1.9 & 1.7 & 1.2 & 7.2 & 82.5 & \cellcolor{gray!25}\textbf{5.5} \\
\end{tabular} \\
\midrule
SAE-opt &
\begin{tabular}{@{}l|cccccc@{}}
 & Rem. & Und. & App. & Ana. & Eva. & Cre. \\ \hline
Rem. & \cellcolor{gray!25}\textbf{10.8} & 14.4 & 0.5 & 18.5 & 38.0 & 17.8 \\
Und. & 1.3 & \cellcolor{gray!25}\textbf{28.87} & 1.6 & 18.9 & 37.23 & 12.1 \\
App. & 1.9 & 19.1 & \cellcolor{gray!25}\textbf{8.40} & 19.50 & 39.1 & 12.0 \\
Ana. & 2.0 & 15.0 & 1.5 & \cellcolor{gray!25}\textbf{31.0} & 43.0 & 7.5 \\
Eva. & 2.1 & 15.92 & 1.0 & 16.13 & \cellcolor{gray!25}\textbf{54.7} & 10.16 \\
Cre. & 2.9 & 18.0 & 1.6 & 16.8 & 42.3 & \cellcolor{gray!25}\textbf{18.4} \\
\end{tabular} &
\begin{tabular}{@{}l|cccccc@{}}
 & Rem. & Und. & App. & Ana. & Eva. & Cre. \\ \hline
Rem. & \cellcolor{gray!25}\textbf{16.5} & 12.7 & 0.5 & 15.5 & 41.0 & 13.8 \\
Und. & 2.8 & \cellcolor{gray!25}\textbf{19.3} & 0.6 & 16.4 & 41.4 & 19.5 \\
App. & 2.7 & 13.6 & \cellcolor{gray!25}\textbf{10.83} & 19.6 & 38.37 & 14.9 \\
Ana. & 2.3 & 12.2 & 0.7 & \cellcolor{gray!25}\textbf{35.4} & 34.6 & 14.8 \\
Eva. & 3.9 & 12 & 0.2 & 17.6 & \cellcolor{gray!25}\textbf{55.8} & 10.5 \\
Cre. & 3.8 & 13.7 & 0.4 & 20.1 & 37.3 & \cellcolor{gray!25}\textbf{24.7} \\
\end{tabular} \\
\bottomrule
\end{tabular}
\end{sc}
\end{table*}

\begin{table}[h!]
\centering
\small
\setlength{\tabcolsep}{4pt}
\caption{Relevance rate (\%) under GPT-4o judge / Claude judge for each (target, method) pair.}
\label{tab:relevance}
\begin{sc}
\footnotesize

\begin{tabular}{lccccccc}
\toprule
Target  & CAA & ITI & SAE-SSV & DiscoQ & Dense-opt& SAE-opt-anch & SAE-opt \\
\midrule
remember  & 71.7 / \textbf{73.3} & 59.4 / 53.7 & 71.0 / 65.2 & 69.0 / 63.8 & 73.0 / 70.0 & \textbf{74.2} / 65.9 & 73.1 / 72.4 \\
understand& 68.7 / 70.9 & 58.1 / 54.4 & 73.1 / 69.3 & 67.7 / 63.8 & 71.7 / 68.0 & 72.4 / 66.5 & \textbf{73.9} / \textbf{71.8} \\
apply     & 71.6 / \textbf{70.1} & 59.9 / 55.1 & 70.2 / 66.4 & 68.7 / 64.0 & 72.6 / 69.1 & 71.5 / 66.2 & \textbf{73.1} / 69.8 \\
analyze   & 69.7 / 72.6 & 52.3 / 51.4 & 72.9 / 71.4 & 67.9 / 64.0 & 72.2 / 69.5 & 72.0 / 66.4 & \textbf{74.4} / \textbf{73.9} \\
evaluate  & 72.3 / 75.9 & 69.2 / 67.6 & 72.8 / 70.2 & 67.3 / 63.8 & 70.5 / 67.1 & 73.1 / 66.6 & \textbf{75.1} / \textbf{76.6} \\
create    & 46.0 / 57.6 & 36.9 / 34.4 & 72.2 / 68.1 & 67.3 / 63.8 & 71.8 / 67.4 & 71.3 / 67.0 & \textbf{73.4} / \textbf{69.0} \\
\bottomrule
\end{tabular}
\end{sc}
\end{table}

 \newpage

\section{Different Layers of Attention}

\label{app:layers}

\begin{figure*}[h!]
    \centering
    \begin{minipage}[t]{0.40\textwidth}
        \centering
        \includegraphics[width=\linewidth, height=6.0cm, keepaspectratio]{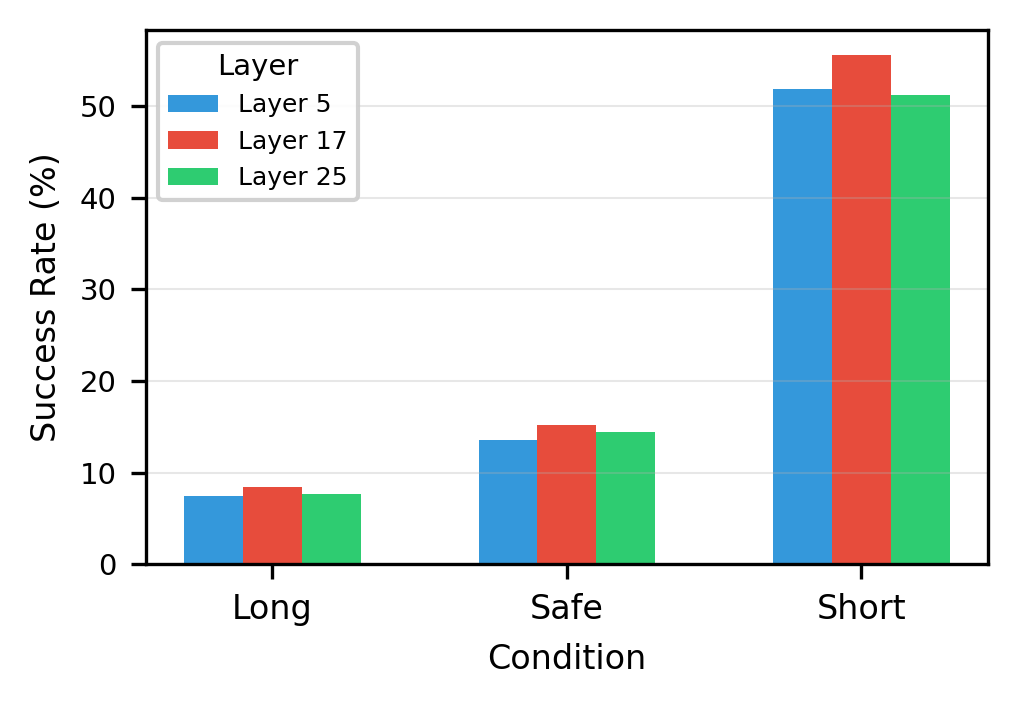}
        \caption{(a) Steering behavior across layers on TGW for Qwen (layers 5, 17, 25).}
        \label{fig:tgw_across_layers}
    \end{minipage}
    \hfill
    \begin{minipage}[t]{0.56\textwidth}
        \centering
        \includegraphics[width=\linewidth, height=6.0cm, keepaspectratio]{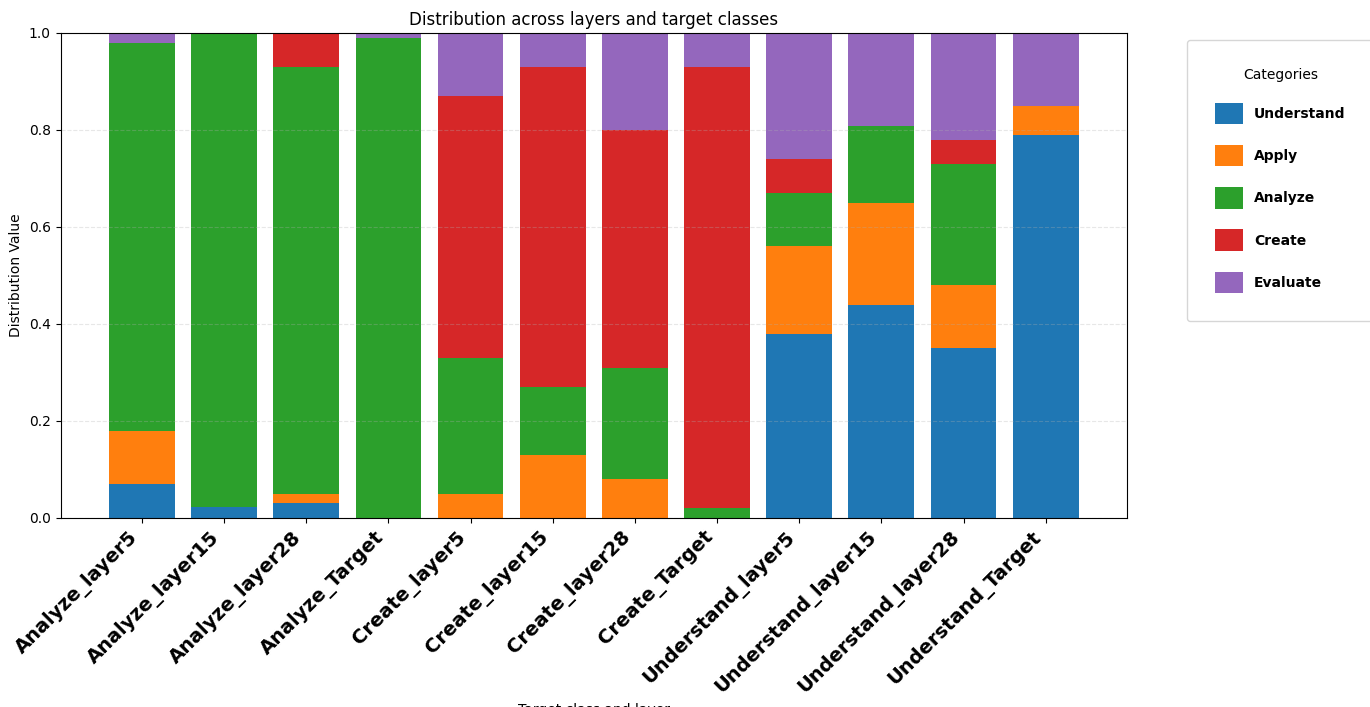}
        \caption{(b) Steering behavior across layers on cognitive style dataset for Phi (layers 5, 15, 28).}
        \label{fig:across_attn_layers}
    \end{minipage}
    \caption{Steering effectiveness across different attention layers. Middle layers consistently yield the strongest and most reliable steering performance in both TGW and cognitive-style tasks.}
    \label{fig:across_layers_combined}
\end{figure*}

We analyze how query embedding sparsity affects steering performance across layers using both TGW (Figure~\ref{fig:tgw_across_layers}) and cognitive style tasks (Figure~\ref{fig:across_attn_layers}), training SAEs at different layers. Across both domains, sparse query features from middle layers consistently yield the most reliable and effective steering, achieving stronger alignment with target attributes than features from early or late layers . This behavior is consistent with prior findings that middle attention layers most strongly represent structured relational dependencies~\cite{vig2019analyzing}.

\section{Steering Entanglement: Detailed Results}
\label{app:entanglement}
Each distance trajectory is normalized by its initial value so all samples begin at 1.0. For a given sample, non-target drift is the absolute difference between the final and initial normalized distance, averaged across non-target classes. The drift ratio reported in the main text is the mean L2 drift divided by the mean L1 drift, averaged across all test samples for each configuration. Table~\ref{tab:entanglement} reports the per-configuration drift values and optimization steps for all models.

\begin{table}[h]
\centering
\begin{sc}
\footnotesize
\caption{Steering entanglement details. \textit{NT Drift}: average absolute change in normalized distance to non-target classes (lower = cleaner)}
\label{tab:entanglement}
\begin{tabular}{llcccccc}
\toprule
\textbf{Model} & \textbf{Target} & \textbf{Coef} & \textbf{L1 Drift $\downarrow$} & \textbf{L2 Drift$\downarrow$} & \textbf{L2/L1} & \textbf{Steps (L1)} & \textbf{Steps (L2)} \\
\midrule
Qwen & short & 0.003 & 0.0423 & 0.090 & $2.13\times$ & 258 & 117 \\
Qwen & short & 0.03 & 0.0237 & 0.116 & $4.9\times$ & 107 & 117 \\
Qwen & safe & 0.003 & 0.055 & 0.082 & $1.49\times$ & 235 & 137 \\
Qwen & safe & 0.03 & 0.0313 & 0.086 & $2.75\times$ & 157 & 133 \\
Qwen & long & 0.003 & 0.0541 & 0.134 & $2.47\times$ & 207 & 147 \\
Qwen & long & 0.03 & 0.0457 & 0.171 & $3.74\times$ & 191 & 139 \\
\midrule
Phi & short & 0.003 & 0.0262 & 0.038 & $1.45\times$ & 136 & 133 \\
Phi & short & 0.03 & 0.0237 & 0.069 & $2.91\times$ & 144 & 135 \\
Phi & safe & 0.003 & 0.0338 & 0.048 & $1.42\times$ & 131 & 128 \\
Phi & safe & 0.03 & 0.0305 & 0.085 & $2.79\times$ & 129 & 128 \\
Phi & long & 0.003 & 0.0279 & 0.042 & $1.5\times$ & 162 & 155 \\
Phi & long & 0.03 & 0.0237 & 0.067 & $2.83\times$ & 162 & 149 \\
\midrule
LLaMA & short & 0.003 & 0.0296 & 0.059 & $1.99\times $ & 407 & 187 \\
LLaMA & short & 0.03 & 0.022 & 0.098 & $4.46\times$ & 179 & 194 \\
LLaMA & safe & 0.003 & 0.0474 & 0.056 & $1.18\times$ & 422 & 219 \\
LLaMA & SAFE & 0.03 & 0.0338 & 0.078 & $2.3\times$ & 273 & 221 \\
lLaMA & long & 0.003 & 0.0406 & 0.100 & $2.46\times$ & 173 & 244 \\
LLaMA & long & 0.03 & 0.0541 & 0.150 & 
$2.77\times$ & 358 & 230 \\

\bottomrule
\end{tabular}
\end{sc}
\end{table}

\end{document}